\documentclass[sn-mathphys-num]{sn-jnl}

\usepackage{hyperref}
\usepackage{graphicx}%
\usepackage{multirow}%
\usepackage{amsmath,amssymb,amsfonts}%
\usepackage[title]{appendix}%
\usepackage{xcolor}%
\usepackage{booktabs}%
\usepackage{algorithm}%
\usepackage[noend]{algpseudocode}
\usepackage{adjustbox}
\usepackage{comment}
\usepackage{pdflscape}

\newlength{\bibitemsep}
\newlength{\bibparskip}
\let\oldthebibliography\thebibliography
\renewcommand\thebibliography[1]{%
  \oldthebibliography{#1}%
  \setlength{\parskip}{\bibitemsep}%
  \setlength{\itemsep}{\bibparskip}%
}

\newcommand{\blind}[1]{#1}

\newcommand{\ale}[1]{{#1}}
\newcommand{\pun}[1]{{#1}}

\newcommand{\sys}{\mbox{\bf TopProRec}}
\DeclareMathOperator{\argmax}{argmax}

\begin{document} 

\title{\ale{Recommending the right academic programs: An interest mining approach using BERTopic}}

\author*[1,2]{\fnm{Alessandro} \sur{Hill}}\email{alessandro.hill@unibo.it}

\author[2]{\fnm{Kalen} \sur{Goo}}\email{kgoo@calpoly.edu}

\author[2]{\fnm{Puneet} \sur{Agarwal}}\email{pagarw05@calpoly.edu}

\affil[1]{\orgdiv{Department of Electrical, Electronic and Information Engineering ``Guglielmo Marconi'' (DEI)}, \orgname{University of Bologna}, \orgaddress{\street{Viale del Risorgimento, 2}, \city{Bologna}, \postcode{40136}, \country{Italy}}}

\affil[2]{\orgdiv{Department of Industrial and Manufacturing Engineering}, \orgname{California Polytechnic State University}, \orgaddress{\street{1 Grand Ave}, \city{San Luis Obispo}, \postcode{93407}, \state{CA}, \country{USA}}}

\abstract{\pun{Prospective students face the challenging task of selecting a university program that will shape their academic and professional careers. For decision-makers and support services, it is often time-consuming and extremely difficult to match personal interests with suitable programs due to the vast and complex catalogue information available. This paper presents the first information system that provides students with efficient recommendations based on both program content and personal preferences. BERTopic, a powerful topic modeling algorithm, is used that leverages text embedding techniques to generate topic representations. It enables us to mine interest topics from all course descriptions, representing the full body of knowledge taught at the institution. Underpinned by the student's individual choice of topics, a shortlist of the most relevant programs is computed through statistical backtracking in the knowledge map, a novel characterization of the program-course relationship. This approach can be applied to a wide range of educational settings, including professional and vocational training. A case study at a post-secondary school with 80 programs and over 5,000 courses shows that the system provides immediate and effective decision support. The presented interest topics are meaningful, leading to positive effects such as serendipity, personalization, and fairness, as revealed by a qualitative study involving 65 students. Over 98\% of users indicated that the recommendations aligned with their interests, and about 94\% stated they would use the tool in the future. Quantitative analysis shows the system can be configured to ensure fairness, achieving 98\% program coverage while maintaining a personalization score of 0.77. These findings suggest that universities could expand student support services by implementing this real-time, user-centered, data-driven system to improve the program selection process.}}

\keywords{Higher education, recommender system, university program selection, machine learning, topic modeling}

\maketitle 

\section{Introduction}

Every prospective college student faces the strategic decision of which program to enroll in. Selecting the right major is crucial, as it serves as a key building block for a successful career, fruitful personal development, and a positive college experience. However, the vast and complex academic program data often makes it very challenging for students to choose the option that best aligns with their interests and expectations. \pun{In 2022, 18.6 million students were enrolled in college programs in the USA \citep{us_enrollment}}, selecting from over 1,800 majors \citep{majors} and 44,000 accredited programs at around 6,000 colleges \citep{us_colleges}. The significant financial investment is evident when considering that 2.5\% of the GDP was spent on higher education in 2020 \citep{higher_education}. \pun{The U.S. ranks second among the Organization for Economic Cooperation and Development (OECD) countries in per-student spending on postsecondary education, investing nearly twice the OECD average \citep{us_oecd}.}

\pun{In 2024, a survey by \citet{students_doubts} revealed that 54\% of college students experienced doubts about their choice of major at least occasionally, even though only 11\% of incoming students are undecided about their career plans \citep{stolzenberg2019american}.} However, more than 50\% of students change their major at least once. For instance, \cite{biggers2008} found that around 35\% of students leaving the computer science (CS) program at the Georgia Institute of Technology (USA) stated that the ``CS program is too narrow and could not accommodate my interests outside of CS.'' A recent study shows that an impressive 44\% of job-seeking graduates regret their major choice \citep{ziprecruiter_survey}.

\ale{Currently, prospective students rely on a variety of resources, including academic advising, university program catalogs, private consulting services, and personal references. We believe that student support could be significantly enhanced by complementing the solutions currently in use. Students often struggle to find the right academic path, primarily due to the challenge of aligning their interests with available programs, which is crucial for academic success \citep{harackiewicz2016interest}. The sheer number of academic programs and courses can be overwhelming. Furthermore, interdisciplinary content, dynamic program structures, and vague high-level program information—such as learning objectives—can be confusing, as they must align with the learning outcomes of all courses within the program \citep{hatzakis2007programme}. We believe these factors largely contribute to the difficulties many students face in identifying a suitable academic program, despite the resources available to them.}

At the same time, educational institutions have a strong interest in enrolling students in well-matched programs. Reducing the number of internal transfers—students who change programs before graduation—can significantly decrease administrative efforts. Additionally, increasing the number of students satisfied with their chosen field of study is likely to have a positive effect on dropout rates and overall student satisfaction, both of which are important metrics for an institution’s reputation. Furthermore, efficient support may reduce the workload for university advising personnel.

In this work, we present a practice-oriented approach to help prospective and transfer students efficiently match their personal interests with available study programs. Our system is based on the central idea of mining large course description data, which represents the body of knowledge for all programs. Using topic modeling, an unsupervised machine learning technique, we search for manageable sets of related keywords, referred to as topics, which can be interpreted as areas of interest. Throughout this work, we use the term interest topics to denote these areas. A comprehensive set of interest topics is presented to students, allowing them to explore and select topics that match their interests. After receiving a student's multi-interest topic input, program relevance scores are calculated using a backtracking method. A final program ranking is then established, listing the most relevant majors for consideration in the ranking.

As we demonstrate in our case study, this approach is capable of both identifying unexpected programs and confirming students' considered fields of study. The decision support provided is instantaneous (automated and efficient), dynamic (utilizing up-to-date program and course data), and comprehensive (matching all available programs). Additionally, the system relies on publicly available program data and Python procedures that incorporate open-source packages. While our case study assumes the selection of a single university, the approach is adaptable to a wide range of educational institutions and can be extended to cover multiple schools simultaneously, both nationally and internationally. We are confident that this new recommendation system can become an important component of future student support.

Moreover, program directors, curriculum committees, and administrative personnel can benefit from this tool in managing programs, leading to increased attractiveness and efficiency. A careful analysis of relevant interest topics and corresponding system recommendations could lead to improved course descriptions, reduced overlap with other programs, enhanced interdisciplinary collaboration, and greater structural synergies.

\pun{In contrast to the recommender systems suggested in the literature, our recommendations are not based on student characteristics such as entrance exam scores or high school performance. Our method has the advantage of being entirely based on open-source data and does not require access to confidential student information.} Furthermore, to the best of our knowledge, no automated method that utilizes course description data or other low-level program content has been proposed thus far.

As highlighted by \citet{tang2024}, there is a growing need for research on diversity, equity, and inclusion. Our system promotes equity by helping individuals who may lack access to conventional support systems—such as lower-income families, first-generation students, and marginalized communities—make informed decisions about their education.

Our work supports both prospective students and universities through the following key contributions:
\begin{itemize}
    \item A novel and broadly applicable machine learning-based decision support system for study program recommendations.
    \item Practical design and implementation guidelines derived from real-world experience at a large university.
    \item Comprehensive qualitative and quantitative evaluation of the information system, based on real recommendations, student surveys, and computational analysis.
    \item \pun{Opportunities to expand existing student support services through a real-time, accessible, user-centered, data-driven system, enhancing the overall program selection process.}
\end{itemize}

The paper is structured as follows. After this introductory section, we review the related literature in Section~\ref{sec:relatedWork}. The proposed information system is described in detail in Section~\ref{sec:approach}, followed by the case study including the system evaluation in Section~\ref{sec:caseStudy}. We conclude by summarizing our findings and providing an outlook on future research in Section~\ref{sec:conclusion}. Supplementary material is provided in the Appendix.

\section{Related Work}\label{sec:relatedWork}

In this section, we review methods employed by past studies to assist students in making informed decisions when selecting the most suitable major at a university. Table \ref{tab:method} presents an overview of the methodologies used to develop corresponding Decision Support Systems (DSS). The methods encompass a range of techniques, including Multiple Criteria Decision Making (MCDM) models, fuzzy systems, Rule-based Method of Knowledge Representation (RBMKR), supervised Machine Learning (ML) algorithms, and unsupervised clustering techniques.

Multiple studies have developed DSS for major selection using MCDM methods where the goal is to identify the best alternative based on multiple decision criteria. \citet{permanasari2020multi} introduced the Hybrid MCDM model tailored for senior high school students in Indonesia, covering majors such as mathematics, natural sciences, social science, and language. This model integrates several MCDM methods, including Simple Additive Weighting (SAW), Technique for Order of Preference by Similarity to Ideal Solution (TOPSIS), and Gray Relational Analysis (GRA), leveraging their respective strengths. The dataset utilized includes information such as test scores, academic reports, and student majors. \citet{khasanah2015fuzzy} developed a DSS using the Fuzzy SAW method to aid high school students in major selection, considering uncertainties in decision-making. Criteria in this model are derived from academic performance, psychological test results, and students' interest questionnaire. Furthermore, \citet{conejero2021applying} applied the TOPSIS method to assess the performance of vocational and educational training programs, emphasizing their impact on employability. The study involved a dataset of 28,000 student records, encompassing both educational and labor-related information.

Prior research has employed multiple reasoning models for major selection, including the rule-based method \citep{al2012prototype} and the Theory of Reasoned Action (TRA) \citep{kumar2013examination}. The former developed a rule-based expert system with an object-oriented database, representing majors as objects with attributes and skill values. Major selection rules are based on essential admission requirements and individual preferences, considering factors like high school scores, graduation year, and grades. On the other hand, \citet{kumar2013examination} utilized the TRA framework to model students' decision-making for business majors, considering attitudes towards job opportunities, social image, personal aptitude, and subjective norms from family, friends, teachers, and counselors. Gender differences and the decided/undecided status of students were also noted as influencing factors in major selection decisions.

Previous studies have explored the application of supervised and unsupervised ML techniques in developing recommender systems for major selection. \citet{zayed2022recommendation} applied supervised ML algorithms, including decision trees, support vector machines, and random forests, to predict optimal majors for individual students. The models were trained on a comprehensive dataset encompassing academic performance, grades, labor market data, salary insights, student experiences, and gender. The random forest model proved most effective, achieving a 97.7\% accuracy. \citet{stein2020college} developed a recommender system using a supervised nearest-neighbor approach for major selection. Utilizing nine years of historical student data, the system based recommendations on students' first two years of courses and performance, evaluating how a student's performance in a major compares to the average for that major.

An unsupervised ML method has been applied by \citet{alghamdi2019fuzzy} and \citet{obeid2018ontology} to provide recommendations. \citet{alghamdi2019fuzzy} devised a fuzzy-based recommendation system considering academic performance and student preferences. They employed a similarity-based clustering technique to group majors based on content or field similarity. Meanwhile, \citet{obeid2018ontology} proposed an ontology-based recommender system focusing on students' skills, interests, and preferences rather than grades. Clusters, formed from graduate student profiles, are utilized to recommend majors based on the similarity of student profiles to the generated clusters.

\begin{table}[htp]
\adjustbox{width=1.0\textwidth}{
\begin{tabular}{cc cc cc cc c} 
\toprule
\multirow{1}{*}{\bf Reference} && \multicolumn{7}{c}{\bf Method}  \\ 
\cmidrule{3-9}
&& MCDM && RBMKR && \multicolumn{3}{c}{\bf ML} \\ 
\cmidrule{7-9}
&& && && Supervised && Unsupervised \\ 
\cmidrule{1-1}\cmidrule{3-3}\cmidrule{5-5}\cmidrule{7-7}\cmidrule{9-9}
\citet{permanasari2020multi} && \checkmark && && && \\
\cmidrule{1-1}\cmidrule{3-3}\cmidrule{5-5}\cmidrule{7-7}\cmidrule{9-9}
\citet{khasanah2015fuzzy} && \checkmark && && && \\
\cmidrule{1-1}\cmidrule{3-3}\cmidrule{5-5}\cmidrule{7-7}\cmidrule{9-9}
\citet{conejero2021applying} && \checkmark && && && \\
\cmidrule{1-1}\cmidrule{3-3}\cmidrule{5-5}\cmidrule{7-7}\cmidrule{9-9}
\citet{al2012prototype}&& && \checkmark && && \\
\cmidrule{1-1}\cmidrule{3-3}\cmidrule{5-5}\cmidrule{7-7}\cmidrule{9-9}
\citet{kumar2013examination} && && \checkmark && && \\
\cmidrule{1-1}\cmidrule{3-3}\cmidrule{5-5}\cmidrule{7-7}\cmidrule{9-9}
\citet{zayed2022recommendation} && && && \checkmark && \\
\cmidrule{1-1}\cmidrule{3-3}\cmidrule{5-5}\cmidrule{7-7}\cmidrule{9-9}
\citet{stein2020college}&& && &&\checkmark && \\
\cmidrule{1-1}\cmidrule{3-3}\cmidrule{5-5}\cmidrule{7-7}\cmidrule{9-9}
\citet{alghamdi2019fuzzy} && && && && \checkmark \\
\cmidrule{1-1}\cmidrule{3-3}\cmidrule{5-5}\cmidrule{7-7}\cmidrule{9-9}
\citet{obeid2018ontology} && && && && \checkmark \\
\cmidrule{1-1}\cmidrule{3-3}\cmidrule{5-5}\cmidrule{7-7}\cmidrule{9-9}
\ale{This Work} && && && && \checkmark \\
\bottomrule
\end{tabular}
}
\caption{Methods used in decision support systems for program selection in the literature.}
\label{tab:method}
\end{table}

Related to program selection, researchers have also developed systems to guide students in selecting the most appropriate courses for given university programs. \citet{al2016automated} created a collaborative recommender system for elective courses, utilizing the $k$-means clustering algorithm to group students based on course grade similarity. The $n$-nearest neighborhood approach was then applied to match students with the most fitting group, and personalized recommendations were generated using association rule mining. \citet{elbadrawy2016domain} took a comprehensive approach, considering factors such as student majors, academic levels, degree requirements, and enrollment patterns. They integrated these features into methods such as matrix factorization, user-based collaborative filtering, and popularity-based ranking, resulting in more accurate predictions compared to existing models.

\citet{jing2017guess} introduced a recommender system for Massive Open Online Courses (MOOCs) based on user interest, demographic profiles, and course prerequisites, demonstrating superior performance on XuetangX, a major Chinese MOOC platform. \citet{bakhshinategh2017course} focused on Graduating Attributes (GAs) in a time-aware course recommendation system, integrating student evaluations to assess courses' impact on personal capabilities and GAs. In a unique approach, \citet{pardos2020designing} introduced a serendipity-focused recommender system aiming to provide students with unexpected yet relevant course recommendations. They evaluated three models, including course2vec and bag-of-words. Additionally, \citet{hatzakis2007programme} utilized a program management framework to enhance curriculum coherence in a Master's-level Information Systems Management course, yielding positive outcomes in both vertical and horizontal coherence according to student surveys.

In this paper, we leverage topic modeling, a statistical unsupervised ML technique, to develop an information system that provides decision support for program selection. Topic modeling helps identify abstract ``topics'' within a collection of documents by analyzing a corpus of text data. It is known to be an essential tool for building effective information systems that involve text analysis \citep{muller2016utilizing}. These topics are assessed based on various metrics such as quality, interpretability, stability, diversity, efficiency, and flexibility \citep{abdelrazek2023topic}. Various topic modeling approaches are discussed in works such as \citet{sharma2017survey} and \citet{jipeng2019short}, which include earlier contributions from \citet{papadimitriou1998latent} and \citet{hofmann1999probabilistic}. These approaches can be broadly categorized into two types: probability-based and embedding-based. Probability-based models use statistical methods to group words together, and some commonly used algorithms are Latent Dirichlet Allocation (LDA) \citep{blei2003latent} and Non-negative Matrix Factorization (NMF) \citep{lee2000algorithms}. However, traditional probability-based methods have drawbacks, such as the need to pre-define the number of clusters and the inability to capture semantic context between words.

To address these challenges, embedding-based methods were introduced, which encode text data to place semantically similar words, sentences, or documents near a cluster's centroid. These techniques often utilize pre-trained embedding models to represent each document as a vector within an embedding space. Notable algorithms in this category include Top2Vec \citep{angelov2020top2vec}, BERTopic \citep{grootendorst2022bertopic}, and Cross-lingual Contextualized Topic Models (CTM) \citep{bianchi2020cross}. Recent advancements involve leveraging deep learning and pre-trained language models, such as BERT \citep{devlin2018bert}, to create Neural Topic Models (NTMs) that offer improved flexibility, effectiveness, and efficiency \citep{zhao2021topic}. Additionally, hybrid approaches have emerged, combining both probabilistic and embedding techniques, like the LDA-BERT model by \citet{basmatkar2022overview}.

Topic modeling offers valuable insights through various visualization techniques, enabling humans to understand and analyze extensive collections of documents \citep{lee2012ivisclustering}. This approach has proven successful in diverse fields, such as bioinformatics \citep{liu2016overview}, marketing \citep{reisenbichler2019topic}, material sciences \citep{rani2021topic}, social media \citep{egger2022topic}, transportation \citep{rose2022application}, and tourism \citep{mishra2021deep}. Moreover, topic modeling serves as a valuable tool within the realm of text mining, where vast amounts of data are extracted and processed, with the aim of uncovering valuable and previously unknown information from web data \citep{johnson2012web}. It can be regarded as a technique for knowledge distillation within text mining, particularly in content mining, which concentrates on generating meaningful clusters of words and documents from unstructured text data \citep{tan1999text}. \citet{foll2021exploring} applied topic modeling to analyze and compare Information Systems (IS) curricula from German universities. Using text mining, the authors efficiently examined over 90 programs and 3,700 modules, identifying key topics such as business intelligence, project management, and programming. This study offers valuable insights for academic institutions, employers, and students, highlighting how curricula align with industry needs while providing a scalable alternative to manual content analysis.

Prior studies have utilized a variety of methodologies to aid students in program or major selection at universities. However, our approach clearly differs from previous work. We focus on interest matching, and leverage unused potential to generate interest topics via data mining. Our system revolves around mining extensive course description datasets. Employing machine learning, we generate manageable sets of interest topics presented to prospective and current students. Upon receiving a student's multi-interest input, program relevance scores are computed using a backtracking method. The final program ranking is then established, listing the top relevant majors for consideration. As demonstrated in our case study, our approach excels in both uncovering unexpected programs and confirming fields already under consideration. In contrast to prevalent recommender systems in the literature, our recommendations do not rely on student features like entrance exam scores or high school performance. Furthermore, to the best of our knowledge, no automated method utilizing course description data has been proposed thus far. 

\pun{Our system is one of the few that uses unsupervised ML to develop a recommender system for program selection, as shown in Table \ref{tab:method}. Unsupervised ML offers several advantages for recommender systems, particularly the elimination of the need for labeled data. This feature makes unsupervised methods more scalable and cost-effective compared to supervised approaches. In contrast to MCDM, which relies on predefined criteria and weights that can introduce bias, unsupervised methods focus solely on patterns observed in the data. This approach enables more adaptive and personalized recommendations, making it especially suitable for scenarios with limited user feedback and diverse or changing preferences.}

\section{The Recommender System}\label{sec:approach}

In this section, we describe the developed decision support system, called Top Program Recommender (\sys). It leverages machine learning, specifically topic modeling—a form of text mining—to extract meaningful information from extensive program data. We provide a detailed description of the modular system, covering the recommendation process, data framework, applied methods, and system interfaces.

\subsection{Overview and Recommendation Process}\label{subsec:OverviewAndRecommendation}

\begin{figure}[htp]\centering
 \includegraphics[width=1\textwidth]{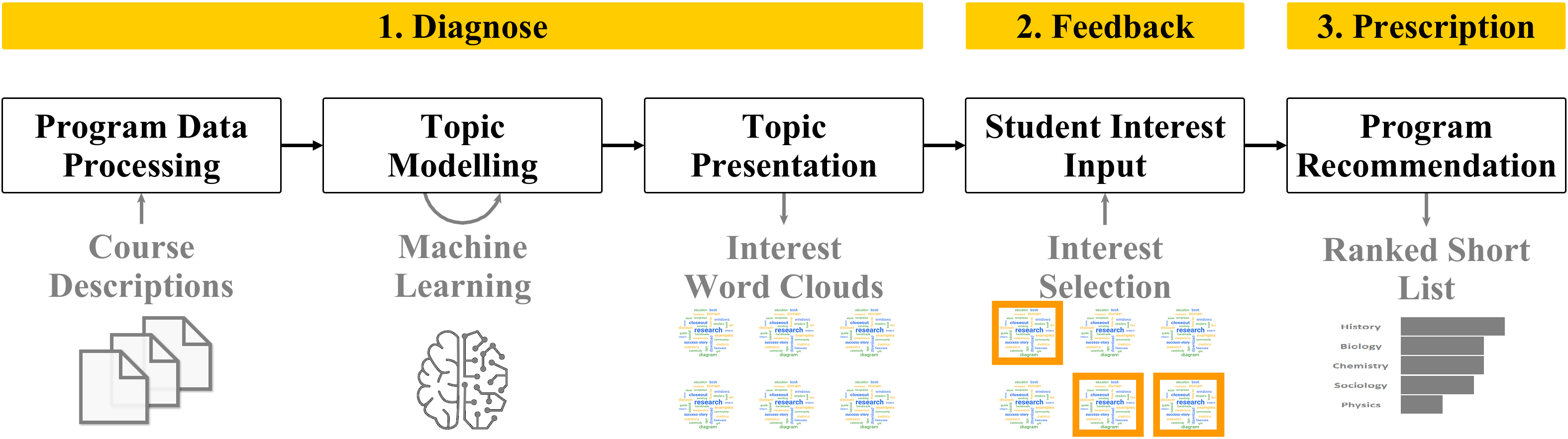}
 \caption{The overall \sys ~decision support process includes the diagnostic phase, the user feedback loop, and the prescriptive analysis. Inputs are course descriptions and study interests; outputs are interest topics and ranked recommended programs.}
 \label{fig:overallProcess}
\end{figure}

Our data-driven approach enables students to quickly receive a curated list of programs that align with their interests. The primary users are prospective students seeking to identify the best program for enrollment. To assist with this, we present a system that suggests a manageable number of finalist programs for further consideration. This system, referred to as \sys, operates as a Recommender System (RS), defined by ~\citet{ricci2022} as ``a software tool that provides suggestions for items most likely of interest to a particular user.'' In our context, the ``overwhelming number of items'' that users must navigate consists of university programs. \sys ~functions as a session-based RS, designed for first-time users without prior user-related data. It generates personalized recommendations based on a single interaction, capturing the user’s interests in real-time. The overall process is illustrated in Figure~\ref{fig:overallProcess}.

The process consists of three main phases: (1) diagnostic analysis of program data, (2) an interactive user feedback step, and (3) the final recommendation of programs. This approach uses both detailed course data and student interests to generate a ranked shortlist of programs. Its prescriptive nature lies in its ability to suggest options for further, in-depth exploration. The core concept used is an `interest topic,' which represents a set of keywords. These keywords, which may appear once or multiple times in course descriptions, are selected by instructors to describe course content. Users can relate these keywords to areas such as study subjects, activities, societal themes, and personal experiences.

In the following sections, we provide a detailed description of the sub-processes. The symbols used are explained in Table \ref{tab:nomenclature}.

{\def\arraystretch{0.85}
\begin{table}[htp]\caption{The symbols, abbreviations and terminology used in different parts of this paper.} \label{tab:nomenclature}
 \begin{tabular}{c c p{0.61\textwidth}}
    \toprule
    & \textbf{Symbol} & \textbf{Explanation}\\
    \midrule
    {\bf Program Data} & $\mathcal{P}$ & Index set for all study programs.\\
    & $\mathcal{C}$ & Index set for all courses.\\
    & $\mathcal{D}$ & Set for all course descriptions consisting of all words; $|\mathcal{D}|=|\mathcal{C}|$.\\
    & $\mathcal{D}^{*}$ & Set for all cleaned course descriptions consisting of keywords only; $\mathcal{D}^{*}_c\subseteq \mathcal{D}_c,~\forall c\in C$.\\
    & $\mathcal{N}$ & Directed knowledge network (map) with node set $\mathcal{C}\cup \mathcal{D}$ and edge set $E$.\\
    & $E$ & Course-program relation $\{(c,p)\in \mathcal{C}\times \mathcal{P}:\textup{course}~c~\textup{can be taken in program}~p\}$.\\
    & $n$ & Number of programs $|\mathcal{P}|$.\\
    & $m$ & Number of courses $|\mathcal{C}|$.\\
    \midrule
    {\bf Topic Modeling} & $\mathcal{T}$ & Index set for computed interest topics.\\
    & $h$ & Number of computed interest topics.\\
    & $\gamma$ & Number of keywords in an interest topic.\\
    \midrule
    {\bf Recommendation} & $\mathcal{R}$ & Index set for recommended programs ($\mathcal{R}\subseteq\mathcal{P}$).\\
    & $\phi$ & Maximal number of interest topics chosen by a user.\\
    & $\tau$ & Number of recommended programs $|\mathcal{R}|$.\\
    & $\textrm{PIS}_p$ & Program interest score for program $p\in\mathcal{P}$.\\
    & $\textrm{R-PIS}_p$ & Relative program interest score for program $p\in\mathcal{P}$.\\
    & $\textrm{SCORE}_p$ & Final program interest score for program $p\in\mathcal{P}$ used for recommendation ranking.\\
    \midrule
    {\bf Evaluation} & $\rho$ & Program reachability (coverage).\\
    \midrule
    {\bf Survey} & $M^{R,P}$ & Binary program recommendation matrix.\\
    & $M^{R,C}$ & Binary college recommendation matrix.\\
    & $\overline{s}$ & Average similarity of recommended items.\\
    \bottomrule
 \end{tabular}
\end{table}
}

\subsection{Program Data Processing}\label{subsec:programDataProcessing}

\begin{figure}[htp]\centering
 \includegraphics[width=1\textwidth]{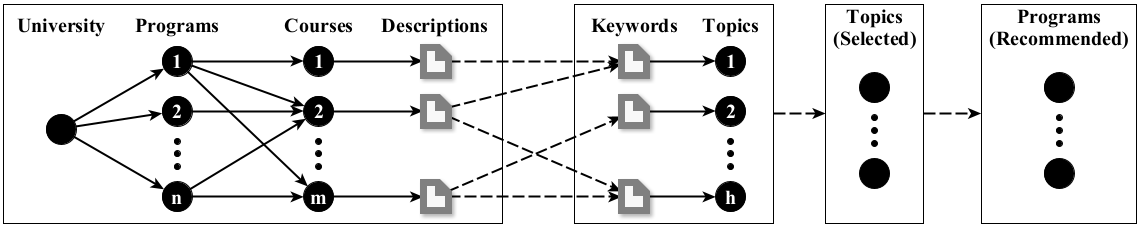}
 \caption{The data scheme used in our framework including $n$ programs and $m$ courses with their descriptions; links (dashed) between descriptions and topics and keywords are computed via machine learning for a given number of desired topics $h$. A subset of user-selected topics is used to calculate the $\tau$ top-recommended programs (right).}
 \label{fig:dataScheme}
\end{figure}

The diagnostic phase begins with the automated extraction of program data which is typically publicly available on the institutional homepage (academic programs, course catalog). This data consists of the list of programs, the list of university courses, and the corresponding course descriptions. 
Let $\mathcal{P}$ denote the index set for programs, $\mathcal{C}$ the index set for courses, and $\mathcal{D}$ the index set for course descriptions, respectively. We illustrate the structure of data elements in Figure~\ref{fig:dataScheme}. A many-to-many relationship $E$ connects courses to programs if they are listed as mandatory (core courses) or optional (elective courses). The bi-modal network $N$ representing this relationship consists of nodes in $\mathcal{C}$ and $\mathcal{P}$ and directed edges in $E$. In Section~\ref{sec:caseStudy} (Figure~\ref{fig:knowledgeMap}), we demonstrate the complexity and scale of the course-to-program relationship using real data. Each course $c\in \mathcal{C}$ is associated with a single course description $D_c$ which is text, commonly restricted to a limited number of words. Note that course names and descriptions are not necessarily unique. Course descriptions serve as input for our topic modeling process, resulting in a set of keywords for each topic, which may not be entirely unique. The data representation shown in Figure~\ref{fig:dataScheme} is important to manage data efficiently but, more importantly, to be able to backtrack from topic keywords to programs, as we will describe later. 

Note that our approach can also be extended to multiple universities, providing recommendations based on an even larger number of potential programs. Conversely, it is also possible to restrict the system to work with only a subset of the institution's programs. 

\subsection{Topic Modeling Method}
Topic modeling is a powerful statistical technique designed to uncover the latent topics or themes present in a collection of documents \citep{vayansky2020review}. Within the realm of topic modeling, a ``word'' serves as the fundamental unit of data. A ``document'' is a sequence of $L$ words, and a ``corpus'' is a comprehensive collection containing $M$ such documents, often representing the entire dataset. In this context, each word represents a course description, a document serves as a collection of course descriptions related to a program, and the corpus encompasses the exhaustive array of programs available for selection. The ``vocabulary'' is the set containing all unique words present in the corpus, while a ``topic'' is represented as a probability distribution that spans this vocabulary. A document can be denoted as $\mathbf{w} = \{w_1, w_2, \dots, w_L\}$, where $w_i$ represents the $i$th word in the sequence and $L$ is the total number of words within that document. Consequently, a corpus is represented by $\mathbf{D} = \{\mathbf{w}_1, \mathbf{w}_2, \dots, \mathbf{w}_M\}$ or more succinctly as $\mathbf{D} = \{d_1, d_2, \dots, d_M\}$, with each $d_n$ equivalent to $\mathbf{w}_n$ signifying the $n$th document in the corpus, and $M$ denoting the total number of documents within the corpus. Topics are represented by various probabilistic or stochastic distributions, depending on the specific method employed. In its simplest form, and when not considering hierarchies or sequential relationships, topics can be represented as $\mathbf{z} = \{z_1, z_2, \dots, z_K\}$, where $z_j$ represents the $j$th topic, and $K$ defines the number of topics that span the entire corpus. The per-topic word distribution is denoted as $\phi(z) = P(w|z)$, representing the likelihood of words given a specific topic $z$. Similarly, the per-document topic distribution is denoted as $\theta(d) = P(z|d)$, representing the likelihood of topics in a given document $d$. The primary goal of topic modeling is to infer the latent variables: the per-document topic distributions $\theta(d)$ and the per-topic word distributions $\phi(z)$. Once the model is trained and the latent variables are estimated, each document can be represented as a mixture of topics, and each topic is represented as a distribution over words. This representation provides valuable insights into the main themes present in the corpus and allows for better understanding and organization of the document collection.

The topic modeling process involves several key steps and elements. It begins with a collection of documents (corpus) comprising text data, such as articles, blog posts, or research papers. Next, the text preprocessing step standardizes the raw text by removing punctuation, converting to lowercase, eliminating stop words, and applying stemming or lemmatization. The feature extraction stage converts the preprocessed text into numerical representations, specifically word frequencies in each document. This transformation enables the data to be utilized by topic modeling algorithms. The core of the process is topic modeling, wherein a topic modeling algorithm (e.g., Latent Dirichlet Allocation (LDA) or Non-negative Matrix Factorization (NMF)) uncovers latent topics present in the collection of documents. These topics are hidden themes or patterns within the data. The output of the topic modeling algorithm provides latent topic distributions for each document in the corpus. Each document is represented as a mixture of topics, where the proportion of each topic indicates its relevance within that document. Additionally, the topic modeling algorithm generates topic-word distributions, which indicate the likelihood of each word occurring within each topic. This information helps to understand the main themes represented by each topic. In this study, we leverage topic-word distributions, also called interest topics in our study, to effectively capture and analyze student interests.

In this paper, we utilize BERTopic \citep{grootendorst2022bertopic}, a powerful topic modeling algorithm that leverages text embedding techniques to generate topic representations. Text embedding techniques encode textual data in a manner that places semantically similar words, sentences, or documents in proximity to a cluster's centroid. For this purpose, BERTopic employs pre-trained language models like all-MiniLM-L6-v2 for English and paraphrase-multilingual-MiniLM-L12-v2, which is applicable to 50 different languages. To improve the clustering algorithm's performance in terms of accuracy and time, a dimensionality reduction technique is incorporated after obtaining the text embeddings. This step reduces the dimensionality of the embeddings, and popular methods such as Principal Component Analysis (PCA), t-distributed Stochastic Neighbor Embedding (t-SNE), or Uniform Manifold Approximation and Projection (UMAP) can be employed for this purpose. Subsequently, the Hierarchical Density-based Spatial Clustering of Applications with Noise (HDBSCAN) algorithm is applied to create clusters from the reduced embeddings. HDBSCAN employs a soft-clustering approach, which allows noise to be modeled as outliers, leading to more robust and accurate clustering results. BERTopic further enhances the topic modeling process by implementing the Class-based Term Frequency-inverse Document Frequency (c-TF-IDF) algorithm \citep{egger2022topic}. This algorithm quantifies the importance of words within each cluster and generates topic-word representations. A higher c-TF-IDF value indicates that a word is more representative of its corresponding topic. BERTopic has demonstrated remarkable success across diverse domains, making it a versatile and effective topic modeling algorithm. Its applications have spanned numerous fields, including tourism \citep{moreno2023examining}, social media \citep{egger2022topic}, healthcare \citep{uncovska2023rating}, and banking \citep{ogunleye2023comparison}.

\begin{figure}[htp]\centering
 \includegraphics[width=1.00\textwidth]{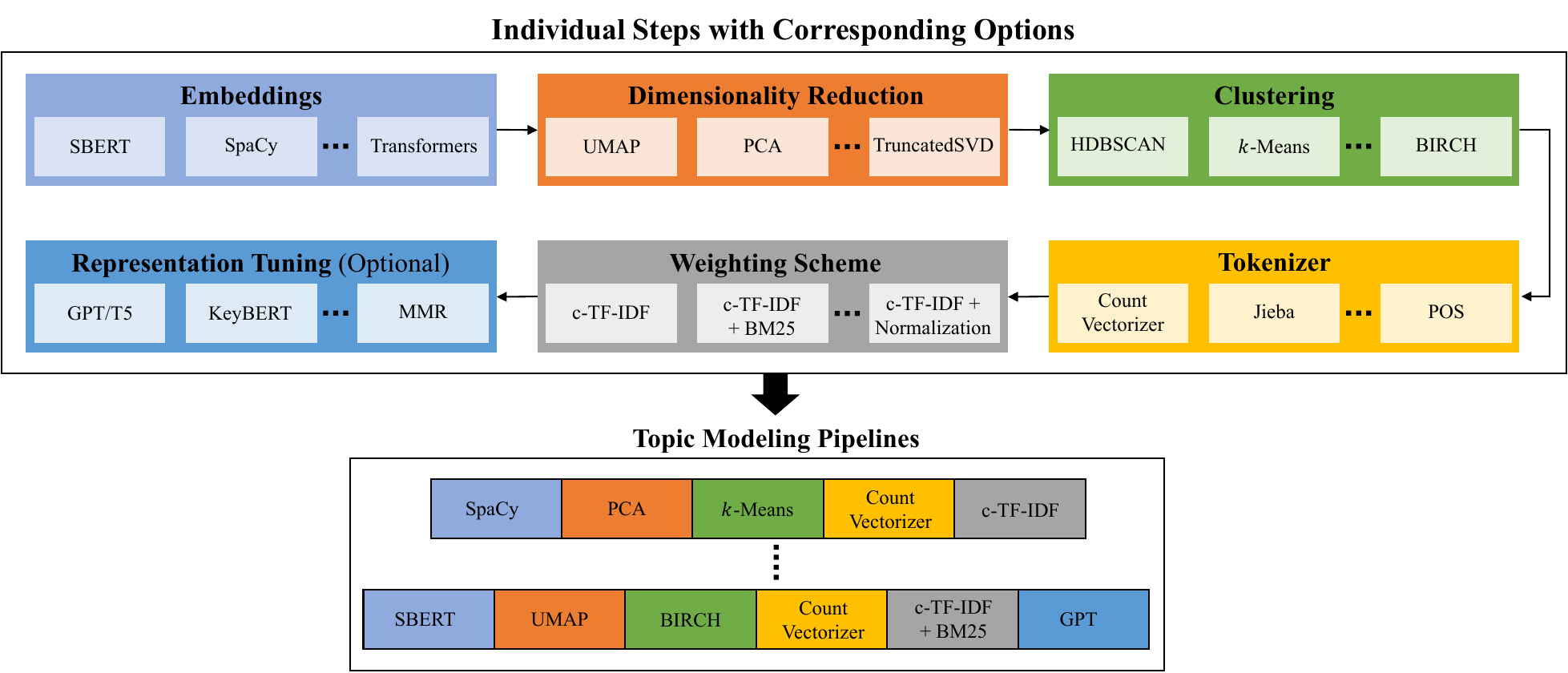}
 \caption{Sequence of steps in creating topic representations with BERTopic.}
 \label{fig:BERT_flowchart}
\end{figure}

Figure \ref{fig:BERT_flowchart} illustrates BERTopic as a sequential process for generating its topic representations. Within each step, various options are available, enabling users to construct a personalized topic modeling pipeline. BERTopic employs the following default models for each step: (a) embeddings: sentence-transformers, (b) dimensionality reduction: UMAP, (c) clustering: HDBSCAN, (d) tokenizer: CountVectorizer, and (e) weighting scheme: c-TF-IDF. Notably, the inclusion of the c-TF-IDF method allows for rapid generation of accurate topic representations. Nonetheless, given the rapidly evolving landscape of Natural Language Processing (NLP), novel methodologies frequently emerge. To stay current with these advancements, the BERTopic framework facilitates the refinement of c-TF-IDF topics by integrating techniques such as GPT, T5, KeyBERT, Spacy, and others.

\pun{BERTopic provides more benefits and strengths in comparison to other topic modeling methods by leveraging contextual embeddings from transformers such as BERT \citep{kenton2019bert}, enabling a deeper understanding of word meanings in context rather than relying solely on word frequency. By combining UMAP for dimensionality reduction and HDBSCAN for flexible clustering, it can effectively handling complex and noisy data structures. The use of c-TF-IDF improves topic interpretability by identifying representative terms for each cluster \citep{grootendorst2022bertopic}. Moreover, its seamless integration with Python libraries such as Hugging Face Transformers \citep{wolf2020transformers} enhances usability and allows for easy incorporation into existing workflows.}

\subsection{Topic Presentation and Interest Feedback}
Once interest topics are computed by the topic modeling pipeline discussed above, we present the topics in order to capture student interests. The computed interest topics are denoted by $\mathcal{T}$. In a pre-processing step, technical content such as class type (e.g., lecture, laboratory), class mode (e.g., virtual, face-to-face), and prerequisite courses are removed. To identify the most relevant keywords for each topic, BERTopic assigns importance scores to individual terms within the topic cluster. Terms that appear frequently within the topic are considered more significant and are more likely to be chosen as keywords. For visual representation, we utilize word clouds (Figure~\ref{fig:overallProcess}) to illustrate each keyword cluster. These word clouds are limited to include only the top $\gamma$ keywords (or terms) based on their importance scores. Words vary in size and color according to their relevance to the respective interest topic as determined by the topic modeling approach. They are placed both horizontally and vertically in a space-optimized layout. Word clouds are presented to the user in a matrix fashion in an arbitrary order. Students are asked to select a subset of up to $\phi$ interest topics after their evaluation with respect to personal interests.

\subsection{Program Recommendation}
Building upon the relationships illustrated in our data scheme (see Figure~\ref{fig:dataScheme}), we developed a network backtracking technique. This method calculates a statistical score for each program based on the selected interest topics (keyword clouds). Specifically, we compute the following two metrics for each program:

\begin{itemize}
    \item Program Interest Score (PIS): We count the number of times that keywords in selected topics appear in one of the courses that can be taken within the program.
    \item Relative Program Interest Score (R-PIS): This relative metric compares PIS to the total number of courses in the program to eliminate the advantage of large programs.
    \item Final score (SCORE): This normalized metric scales the R-PIS to the interval [0,100] and is finally presented to the user together with the recommendation.
\end{itemize}
We present the $\tau$ top programs ($\mathcal{R}$) as a ranked list to the student, highlighting programs worth considering in greater detail. For the final decision-making process, students are encouraged to consult resources specific to each program, course, and department.

The complete backtracking method including the metric calculation is described in Algorithm~\ref{alg:backtracking}. First, the three interest scores (PIS, R-PIS, SCORE) are initialized in lines 1-4. We assume that the user selected at least one interest topic; i.e., $\mathcal{T}\neq\emptyset$. In the actual backtracking, we iterate over the user-selected interest topics (line 5) and the corresponding keywords (line 6). For every course, we check if the considered keyword is contained in the cleaned course description (lines 7-8). If this is the case then we increment the PIS for each program that the course is associated with (lines 9-10). Note that this can be done by using the arcs $E$ in the course-program network $N$ (see Section~\ref{subsec:programDataProcessing}). Using the programs' PIS values, we calculate the R-PIS next (lines 11-12). Then, at most $\tau$ top programs ($\mathcal{R}$) are identified by sorting the programs in descending order with respect to PIS (lines 13-16). We do not recommend programs that have no interest overlap; i.e., if the R-PIS value equals to zero. The SCORE values are calculated for the recommended programs (lines 17-18) before the procedure returns them together with all metric values (line 19).

\makeatletter
\newcommand{\algrule}[1][.2pt]{\par\vskip.5\baselineskip\hrule height #1\par\vskip.5\baselineskip}
\makeatother

\begin{algorithm}
\caption{Backtracking Method (Input: User-Selected Interest Topics $\mathcal{T}$)}
\label{alg:backtracking}
\begin{algorithmic}[1]
\For{$p \in \mathcal{P}$}\Comment{Initialize program interest scores.}
 \State $\textrm{PIS}_p\leftarrow 0$
 \State $\textrm{R-PIS}_p\leftarrow 0$
 \State $\textrm{SCORE}_p\leftarrow 0$ 
\EndFor
\algrule
\For{$T \in \mathcal{T}$} \Comment{Calculate program interest score via backtracking.}
 \For{$w \in T$} 
  \For{$c \in \mathcal{C}$} 
   \If{$w\in d^{*}_c$} 
    \For{$(c,p) \in E$} 
     \State $\textrm{PIS}_p \leftarrow \textrm{PIS}_p + 1$
    \EndFor
   \EndIf  
  \EndFor
 \EndFor
\EndFor
\algrule
\For{$p \in \mathcal{P}$} \Comment{Calculate relative program interest score.}
 \State $\textrm{R-PIS}_{p}\leftarrow \textrm{PIS}_p/|p|$
\EndFor
\algrule
\State $\mathcal{R}\leftarrow \emptyset$ \Comment{Identify top recommended programs.}
\Repeat
 \State $\mathcal{R} \leftarrow \mathcal{R} \cup \{\argmax_{p\in\mathcal{P}\setminus\mathcal{R}}\textrm{R-PIS}_p\}$
\Until{$\argmax_{p\in\mathcal{P}\setminus\mathcal{R}}\textrm{R-PIS}_p = 0 \vee |\mathcal{R}|=\tau$}
\algrule
\For{$p \in \mathcal{R}$} \Comment{Calculate final program interest score.}
 \State $\textrm{SCORE}_{p}\leftarrow 100\cdot\textrm{R-PIS}_p/(\max_{p\in\mathcal{R}}{\textrm{R-PIS}_p})$
\EndFor
\algrule
\State \Return $(\mathcal{R},\textrm{PIS},\textrm{R-PIS},\textrm{SCORE})$ \Comment{Return recommended programs and interest scores.}
\end{algorithmic}
\end{algorithm}

\section{Case Study}\label{sec:caseStudy}

We conducted a comprehensive case study in which we developed a fully functional prototype of the \sys ~system. The system was based on real-world data for the academic year 2021/22 from a large polytechnic university in the USA. The following sections provide a detailed overview of the development process and system architecture, followed by a rigorous experimental evaluation with 65 users, approved by the university's Institutional Review Board (IRB).

\subsection{Data Collection}
We utilized data from 84 programs\footnote{\blind{\scriptsize\url{catalog.calpoly.edu/programsaz}}} across six colleges, encompassing all majors, interdisciplinary minors, and concentrations within the business school. The course data\footnote{\blind{\scriptsize\url{catalog.calpoly.edu/coursesaz}}} included 4,251 course descriptions, each containing an average of 42.3 words, with a maximum allowed length of 50 words per description. In total, we extracted 19,283 keywords, averaging 16.8 characters in length and composed of approximately 2.0 individual words per keyword. The number of courses per program ranged from 11 to 224, with an average of 75.1 courses per program. Courses were associated with at least one program and up to 22 programs—as in the case of Calculus IV—averaging 2.1 programs per course. The university has a total enrollment of about 21,000 students, with approximately 6,000 students enrolled in the largest college, namely Engineering.

\begin{figure}[htp]\centering
     \blind{
     \includegraphics[width=1.0\textwidth]{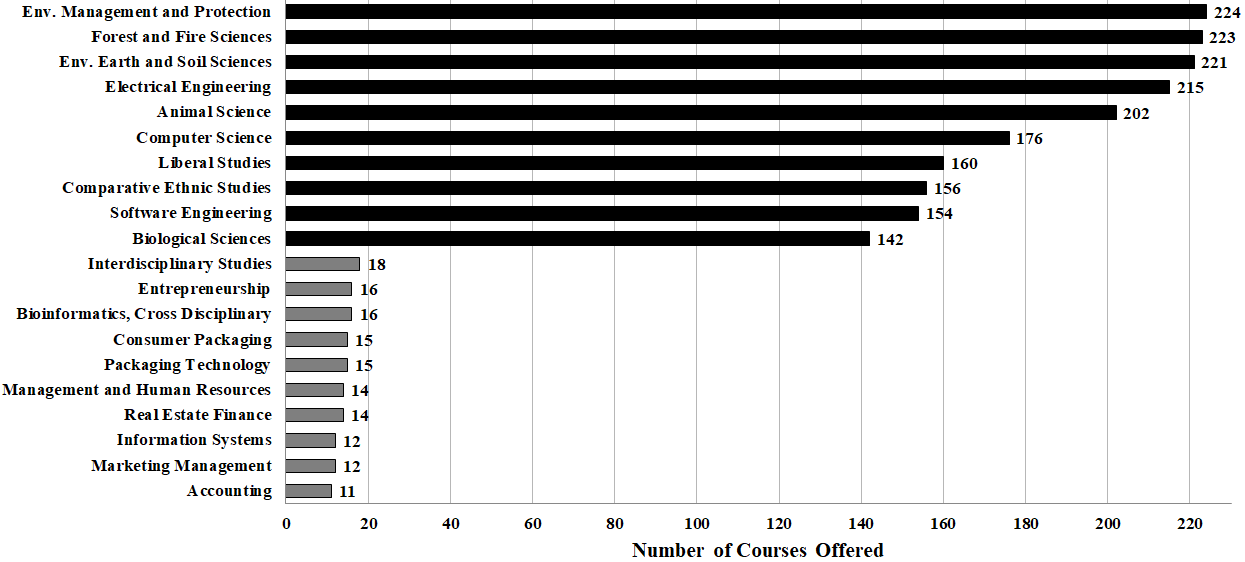}
     }
 \caption{The number of credit courses per program for the top and bottom 10 programs.}
 \label{fig:CoursesPerProgram}
\end{figure}

The number of courses offered by each college is provided in Table \ref{tab:CoursesPerCollege}. Note that Interdisciplinary Programs only contains one program: \blind{Liberal Arts and Engineering Studies}.
\begin{table}[htp]
\caption{The six colleges and interdisciplinary studies with the number of courses that they offer.}\label{tab:CoursesPerCollege}
  \begin{tabular}{c l c c c l c}
   \toprule
   {\bf } & {\bf College} & {\bf \#} && {\bf } & {\bf College} & {\bf \#}\\
   \cmidrule{1-3}\cmidrule{5-7}
    1 & \blind{Agriculture, Food \& Environ. Sciences} & 1768 && 5 &  \blind{Architecture \& Environ. Design} & 328 \\
	2 & \blind{Engineering} & 1446 && 6 & \blind{College of Business} & 286 \\
	3 & \blind{Liberal Arts}	& 1302 && 7 & \blind{Interdisciplinary Programs} & 82 \\
	4 & \blind{Science \& Mathematics} & 1098 && & & \\
   \bottomrule
  \end{tabular}
\end{table}
The number of courses that can be taken in a program varied significantly, ranging from 224 (\blind{Environmental Management and Protection}) to 11 (\blind{Accounting}). Figure \ref{fig:CoursesPerProgram} shows the top 10 and bottom 10 programs based on the number of eligible courses they offer. On average, there were 92.9 courses per program.

\begin{figure}[htp]
\includegraphics[width=1.0\textwidth]{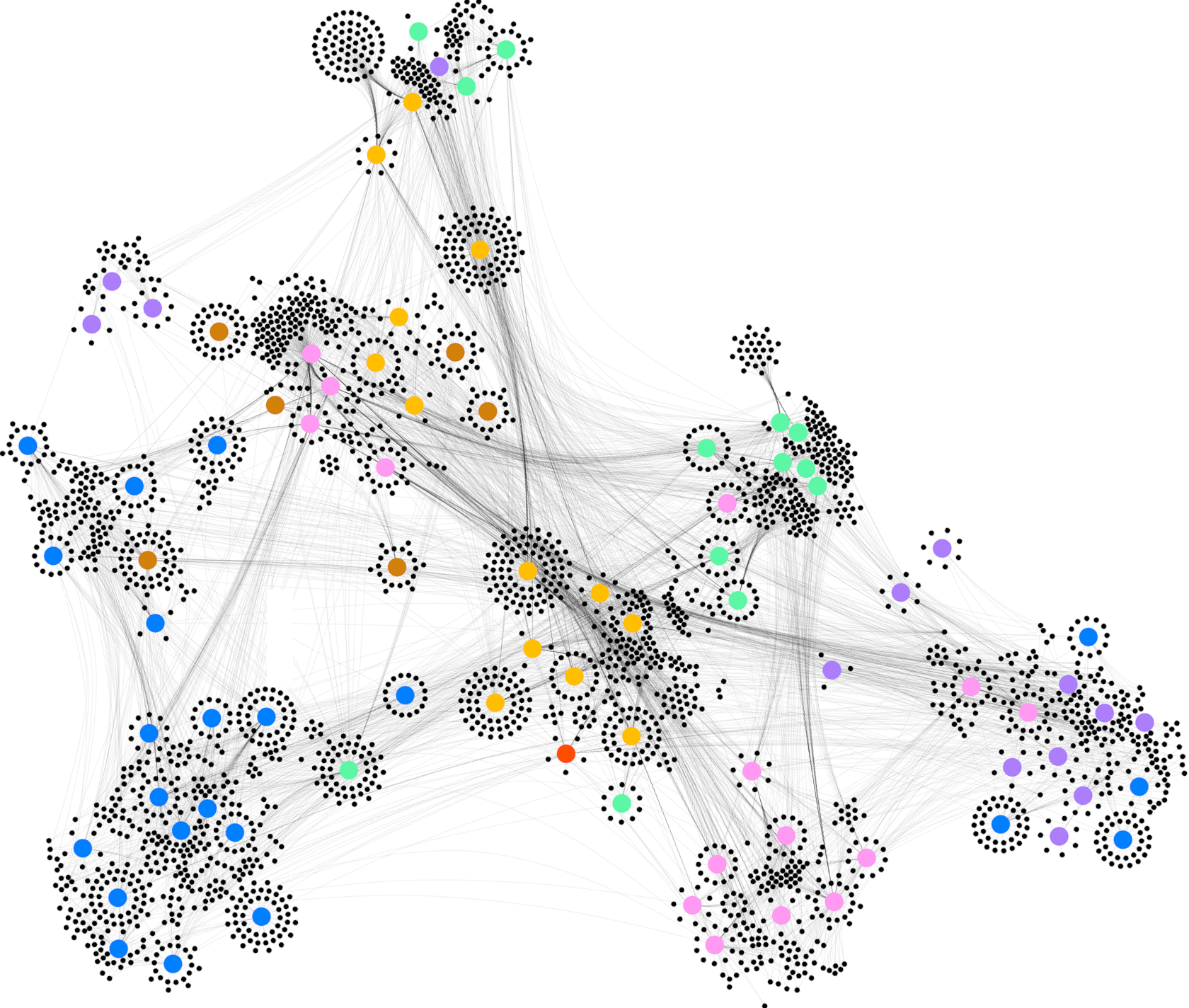}
 \caption{\vspace{-0mm}The ``knowledge map'' representing the relationship ($E$) between courses (2565 smaller black nodes) and programs (84 larger nodes colored by college) at the university; 6143 links indicate that courses are mandatory or optional in programs.}
 \label{fig:knowledgeMap}
\end{figure}

Additionally, we analyzed the corresponding network ($N$) with two categories of nodes (programs, courses) and edges if courses can be taken in programs. We call this network the {\it knowledge map}. This complex relationship between programs and courses is illustrated in Figure~\ref{fig:knowledgeMap}. Larger nodes represent programs, color-coded depending on the college and small black nodes are associated with individual courses. The organic layout is the optimized result of simulating physical forces and using the Louvain modularity method as the natural clustering algorithm. We used the Python package NetworkX\footnote{\url{networkx.org}} to construct and represent the network and the free network diagramming software, yEd\footnote{\url{yworks.com/products/yed}}, to generate the layout and visualize it.

\subsection{System Architecture and Interfaces}

\begin{figure}[htp]\centering
 \includegraphics[width=0.85\textwidth]{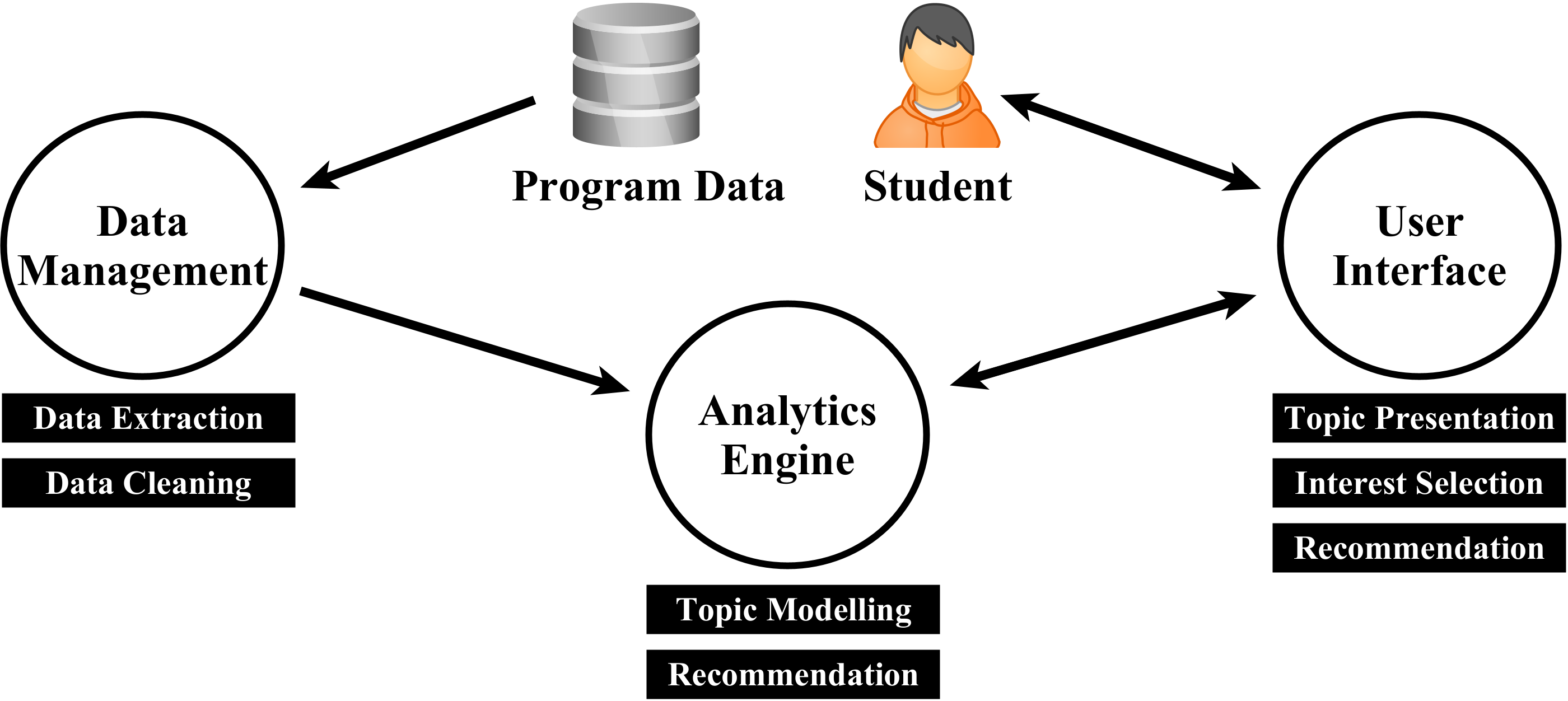}
 \caption{System architecture for the \sys ~recommender system and the main processes at each entity.}
 \label{fig:systemArchitecture}
\end{figure}

We suggest the implementation of \sys ~using a modular system architecture. The system design shown in Figure \ref{fig:systemArchitecture} and explains the interplay of the individual modules including, program data sources, data management, analytics engine, and user interface.

\begin{figure}[htp]\centering
 \includegraphics[width=0.85\textwidth]{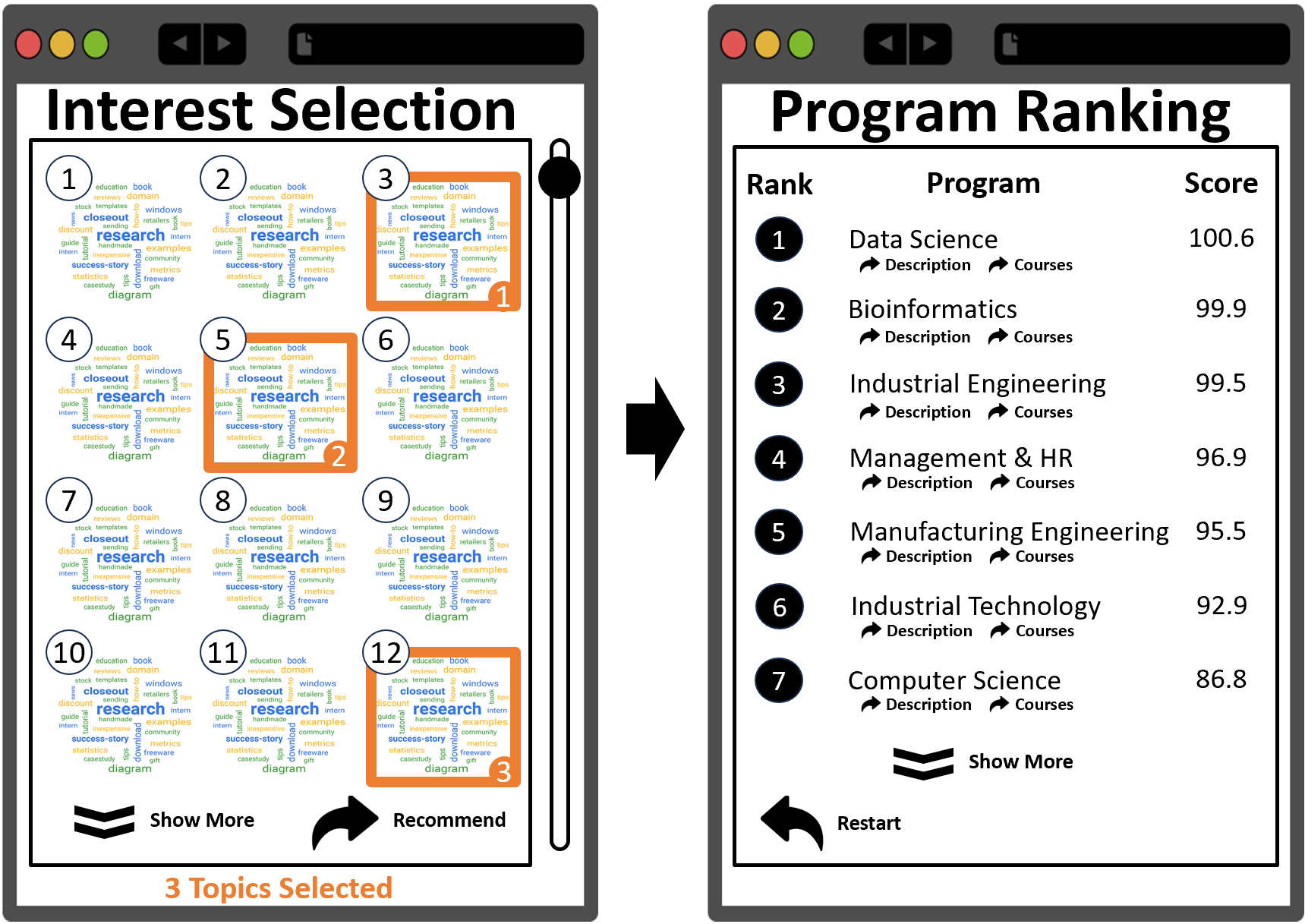}
 \caption{User interface for the program recommendation system including interest topic selection screen (left) and program recommendation screen (right).}
 \label{fig:interfaceMockup}
\end{figure}

Figure \ref{fig:interfaceMockup} shows a \sys ~interface that could be used on personal computers as well on mobile devices. On a first screen, all computed interest topics are presented to the user (left). After the user selects a subset of topics that match personal interests, the computation of the recommendation (Algorithm \ref{alg:backtracking}) is triggered. The recommended programs are then displayed on a second screen together with the final scores (right). Moreover, the user is offered the option to restart the process with a revised interest topic selection.

\subsection{Implementation}\label{subsec:implementation}
We implemented the system using Python 3.10 on a MacOS machine with a 2.6 GHz 6-core Intel Core i7 processor and 16 GB of RAM. For the steps in the process described in Section \ref{subsec:OverviewAndRecommendation}, we used the following packages. The online program and course data retrieval was done using a Python library called Beautiful Soup \citep{beautiful_soup} that provides a convenient way to parse and navigate the content of web pages, making it easier to extract specific information such as text, links, and data tables. We utilized the Natural Language Toolkit (NLTK) \citep{bird2009natural} to preprocess text data using functionalities such as tokenization, stop word removal, stemming, and lemmatization. These preprocessing steps are crucial before feeding text data into a topic modeling algorithm as they help reduce noise and improve the quality of topics generated. Topic modeling was performed via the BERTopic \citep{grootendorst2022bertopic} implementation in version v0.13.0, together with the built-in CountVectorizer from scikit-learn \citep{scikit-learn}. CountVectorizer is a text processing technique that converts text data into a numerical format, making it suitable as input for machine learning algorithms. Word clouds were visualized by WordCloud \citep{word_cloud}.
Furthermore, we used the following system parameter values (see Section~\ref{sec:approach}): $h=30$ (Number of interest topics); $\gamma=20$ (Number of top keywords per interest topic); $\phi=8$ (Maximum number of user-selected interest topics); $\tau=7$ (Number of top programs in final ranking).

\subsection{A Real Recommendation}\label{subsec:realRecommendation}

\begin{figure}[htp]\centering   
 \includegraphics[width=0.98\textwidth]{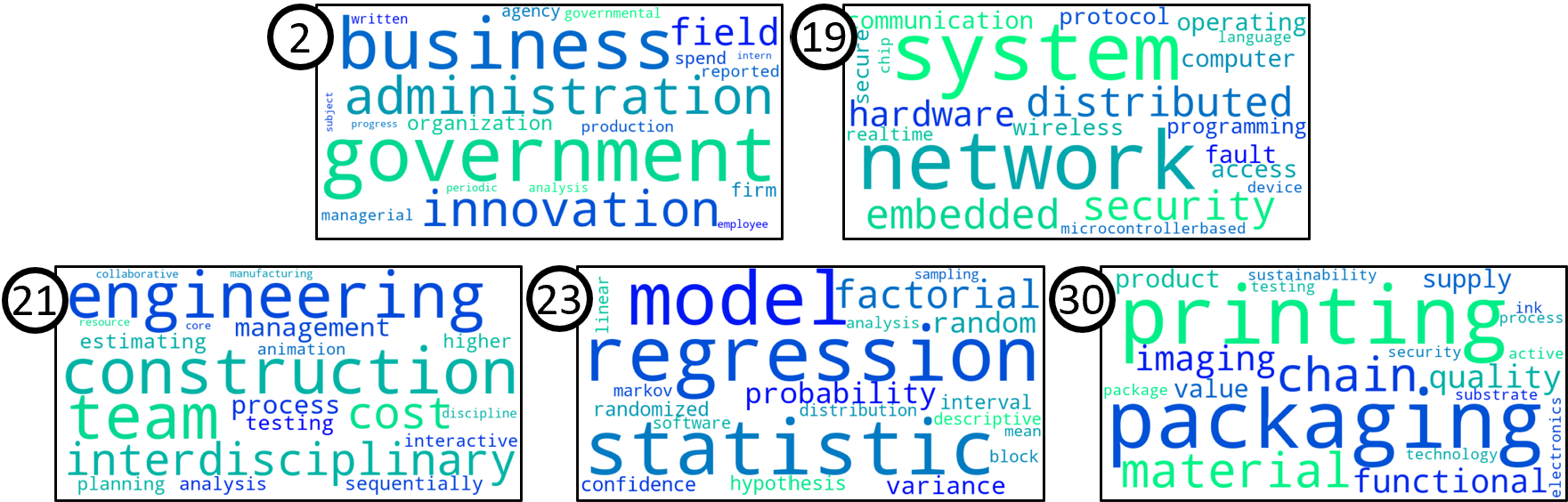}
 \caption{The five (out of 30) user-selected interest topics (word clouds): \#2, \#19, \#21, \#23, \#30.}
 \label{fig:exampletopicwordclouds}
\end{figure}

In the following, we illustrate our system by providing a full example of program recommendation based on the actual data and a fictitious student. Let the 5 word clouds (\#1, \#18, \#20, \#22, \#29) in Figure~\ref{fig:exampletopicwordclouds} correspond to the selection made by the student. We observe that the presented topics are related and the programs can be associated with the keywords contained in the corresponding topics. The remaining 25 unselected word clouds can be found in Appendix A.
For this selection, our backtracking algorithm computed the metrics and ranking shown in Table~\ref{tab:exampleRanking}. Although the PIS scores significantly vary for the presented programs, the R-PIS scores are consistently high. Solely Computer Science, which is ranked last, is shows a lower R-PIS and final score, after a 7-point drop. Now it is up to the student to further examine the programs in the list before making a final decision.   
\begin{table}[htp]
\ale{
\adjustbox{width=\textwidth}{
\begin{tabular}{cc lc cc cc cc cc}
 \toprule
 {\bf Rank} && {\bf Program} && {\bf \#Courses} && {\bf PIS} && {\bf R-PIS} && {\bf SCORE}\\
 \cmidrule{1-1}\cmidrule{3-5}\cmidrule{7-11}
 1 && Data Science && 21 && 113 && 5.381 && 100.0 \\ 
 2 && Bioinformatics && 16 && 86 && 5.375 && 99.9 \\  
 3 && Industrial Engineering && 102 && 546 && 5.353 && 99.5 \\
 4 && Management \& Human Res. && 14 && 73 && 5.214 && 96.9 \\
 5 && Manufacturing Engineering && 115 && 585 && 5.087 && 94.5 \\
 6 && Industrial Technology && 27 && 135 && 5.000 && 92.9 \\
 7 && Computer Science && 176 && 822 && 4.670 && 86.8 \\
 \bottomrule
\end{tabular}
}
\caption{The ranking (top 7) of the recommended programs based on the student interests using the final score (SCORE); program interest score (PIS) and relative program interest score (R-PIS).}\label{tab:exampleRanking}
}
\end{table}

\subsection{System Evaluation}\label{subsec:validation}

In this section, we evaluate \sys~using common metrics typically applied to recommender systems. Our primary goal is to determine whether our tool can effectively support both prospective and current students in practice. The first part focuses on a quantitative analysis of the system design, while the second part presents a qualitative analysis based on a survey. Note that both aspects rely on the program data used and the student population considered. We begin by describing the experimental setup used to gather system performance data.

\subsubsection{User Experiment}
To capture the usefulness of our tool in practice, we conducted an experiment in which each participant went through the full recommendation process. After selecting interests, the personalized program ranking was presented to the user. In a survey, participants were asked about their experience. We randomly selected 64 students from 39 different majors, of which 50 were students at the experiment university and 14 from other local colleges. They all participated voluntarily and no incentive was provided. Students were in there first year (5), second year (22), third year (13), fourth year (14), and beyond (11). We configured our system to $h=30$ interest topics containing $\gamma=20$ keywords each. Users were asked to select up to 8 interest topics to match their interests. The list of recommended programs ranked the top $\tau=7$ programs. How often a program was recommended to a user and how often at what rank is shown in Appendix B.
Note that overall 63 programs were recommended to the 65 users. Similarly, the corresponding rank and recommendation analysis for the colleges is given in Appendix B.
Interdisciplinary Programs was not recommended at all. This is probably due to the fact that it only contains one program. We will show in our evaluation (Section \ref{subsubsec:SystemDesignEval}) that theoretically 98.8\% (83 out of 84 programs) can appear in the recommendation for the parameters used. Solely the program ``Plant Sciences'' is excluded. 

\subsubsection{System Design}\label{subsubsec:SystemDesignEval}

In the following, we evaluate the system design measuring explainability, user controllability, fairness, robustness, bias, privacy protection, personalization, and trustworthiness.
\paragraph{Explainability}
\ale{Explainability focuses on making the recommendation process and the reasons behind specific recommendations clearer to the users \citep{Tsai2020TheEO}. \ale{To increase explainability, we suggest a system add-on that provides information on the importance of user-selected interest topics for the recommended programs. For a topic $t\in\mathcal{T}$ and a program $p\in\mathcal{P}$, let the topic score be defined as the normalized relative number of appearances of keywords related to $t$ in the course descriptions of $p$. Using this information, we can explain to the user which topic choice was least or most influential in the recommendation of a program. This insight can help the user revise their selection of interest topics and experiment in a sensitivity-analysis fashion. The table in Figure \ref{fig:exampletopicscores} shows the corresponding values for the example topic selection and the recommended programs. Additionally, the topic scores for a runner-up program (Statistics; SCORE = 0.859), an alternative engineering program (Civil Engineering; SCORE = 0.481), and a liberal arts program (History; SCORE = 0.136) are presented. As expected, non-recommended programs show low scores for most topics. Interestingly, Topic 19 is not represented in any of the courses of the recommended program Management and Human Resources, and Topic 30 is largely responsible for the recommendation of the Industrial Technology program. This information can also help users understand why they are presented with unexpected programs.}
\begin{figure}[htp]\centering   
 \includegraphics[width=0.98\textwidth]{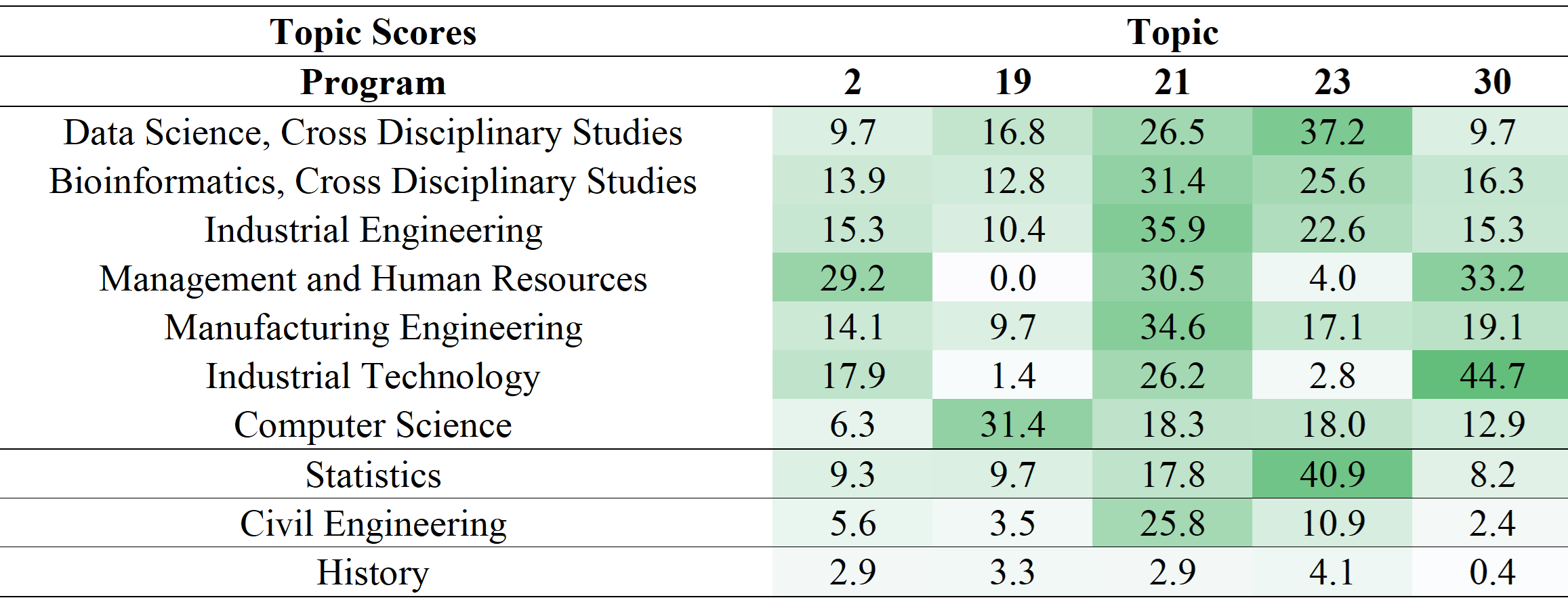}
 \caption{\ale{Topic scores for the user-selected interest topics (\#2, \#19, \#21, \#23, \#30), the 7 recommended programs and 3 programs that were not recommended (Statistics, Civil Engineering, History).}}
 \label{fig:exampletopicscores}
\end{figure}
\noindent
Furthermore, we believe that a degree of explainability is achieved by providing each user with an introductory guide that explains the basic principles and process. For a more detailed discussion on implementing explainable recommender systems, we refer to \citet{tintarev2015explaining}.}

\paragraph{User Controllability}
\sys ~does not allow the user to customize and participate in the recommendation method. However, such user controllability~\citep{Tsai2020TheEO} could be achieved by giving the user access to parameters such as the number of presented interest topics ($h$), or allowing the application of input filters on the list of programs and colleges (e.g., only show engineering programs). \ale{Depending on the education system and the institution, it can be very important to consider program requirements such as minimum entry scores. Our system can be extended with user-specified filters that exclude non-eligible programs. Note that the interest topics need to be recomputed to avoid considering irrelevant course descriptions. On a parallel note, a useful extension could allow users to include certain programs a priori. Using the topic scores (Figure~\ref{fig:exampletopicscores}), users could then understand why a current or favored program was not recommended.}

\begin{figure}[htp]\centering
 \includegraphics[width=1\textwidth]{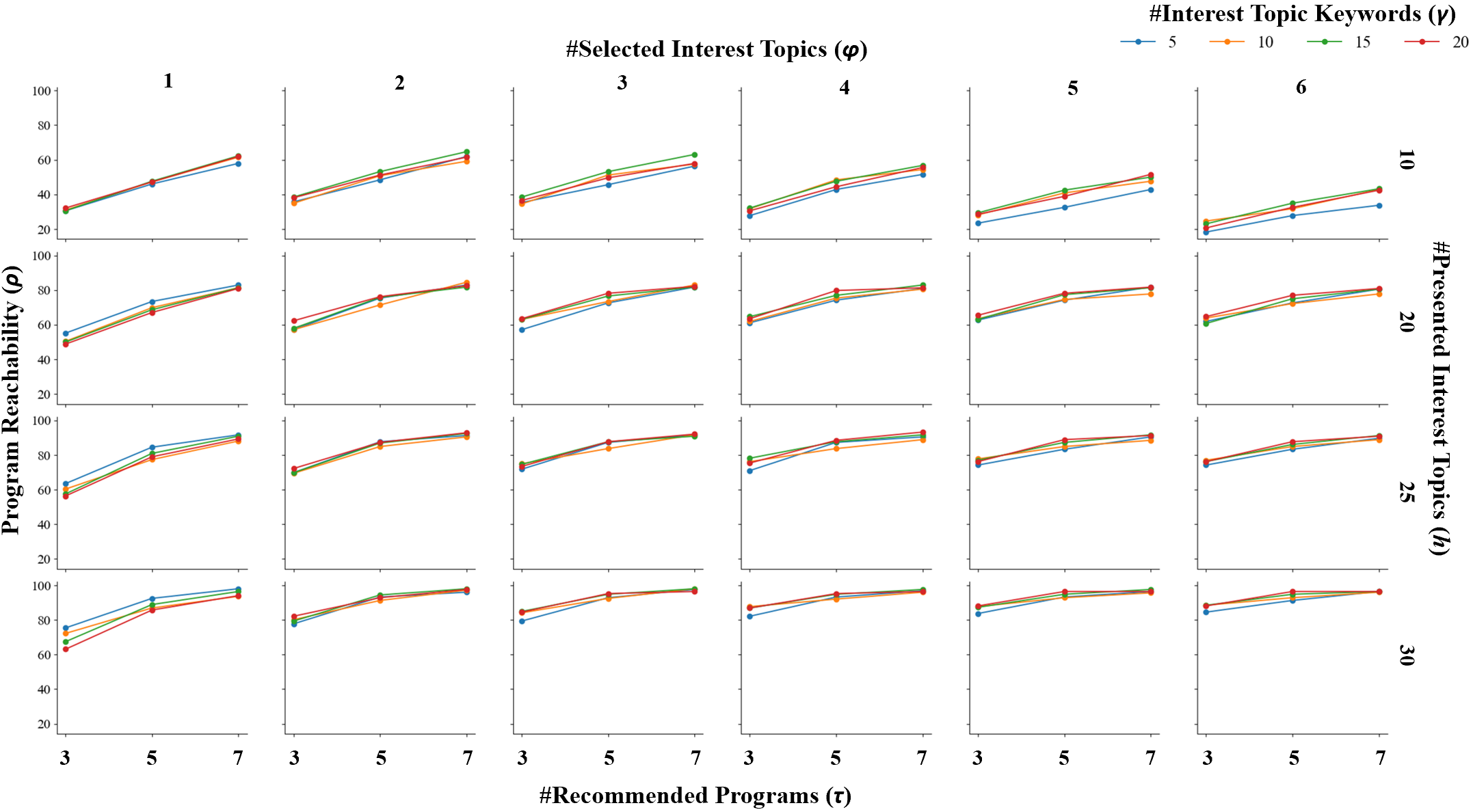}
 \caption{Sensitivity analysis on program reachability ($\rho$) for various parameter configurations ($h\in\{10,20,25,30\}$: \# interest topics; $\tau\in\{3,5,7\}$: \# recommended programs; $\gamma\in\{5,10,15,20\}$: \# of topic keywords; $\phi\in\{1,\ldots,6\}$: \# selected interest topics).}
 \label{fig:programReachability}
\end{figure}

\paragraph{Fairness}
A well-known issue regarding the fairness is the under-recommendation of items \citep{li2023fairness,pitoura2022fairness,deldjoo2023fairness,Ekstrand2022}. Programs that are under-ranked could seriously suffer from not being considered by future students. It is not immediately clear whether our tool is equitable with respect to the programs offered by the university. That is, does the model allow the declaration of interests such that potentially every program is suggested? Therefore, we investigate the program reachability; i.e., the programs that potentially appear in the final ranking based on the user input. In the context of recommender systems, this number of items that can be recommended for a given training data is also called {\it coverage}. In order to better understand the impact of parameters $h$ (\# topics), $\phi$ (\# selected topics), $\gamma$ (\# words in word cloud), and $\tau$ (\# programs in ranking), we conduct a sensitivity analysis. For $h\in\{10,20,25,30\}$, selected interest topics $\phi\in\{1,...,6\}$, number of recommended programs $\tau\in\{3,5,7\}$, and number of topic keywords $\gamma\in\{5,10,15,20\}$, we calculate all possible rankings using all possible combinations of $\phi$ topics. We define program reachability $\rho$ as the relative number of programs in $\mathcal{P}$ that were suggested in the top-$\tau$ ranking at least once. That is, for a given parameter quadruplet $(\tau,\phi,\gamma,h)$:
$$\rho=100\cdot \frac{\big|\bigcup_{T\subset \mathcal{T},|T|=\phi}\mathcal{R}(T,\phi,\gamma,h)\big|}{|\mathcal{P}|}$$
Here, $\mathcal{R}(T,\phi,\gamma,h)$ denotes the set of recommended programs obtained by \sys ~when using the corresponding parameter values. The results are shown in Figure~\ref{fig:programReachability}.

We observe that the reachability $\rho$ ranged from 16.7\% (14 programs; $h=10,\phi=6,\gamma=15,\tau=3$) to 100.0\% (84 programs; $h=30,\phi=1,\gamma=5,\tau=7$). A reduced number of presented interest topics ($h$) led to poor reachability. For example, when setting $h=10$, $\rho$ was at most 66.7\%. The number of user-selected interest topics ($\phi$) had an ambiguous impact on the reachability. For higher values of $h$, increasing $\phi$ results in a better $\rho$. Conversely, $\rho$ decreased when working with lower values of $h$. It is not clear how the number of keywords ($\gamma$) in the interest topics influences $\rho$. Although it seems to have a minor impact, augmenting $\gamma$ resulted in both improved reachability (e.g., $\phi=2,h=25$) and deteriorated reachability (e.g., $\phi=1,h=30$). Based on these numbers, we decided the parameter setting in our case study (Section~\ref{sec:caseStudy}). We suggest a configuration that yields a reachability of $\rho>90\%$. This can be achieved by using parameters $h\in\{25,\ldots,30\}$, $\phi\in\{3,\ldots,6\}$, $\gamma\in\{5,\ldots,20\}$, and $\tau\in\{5,6,7\}$. The observed coverage values are high compared to what is often seen with collaborative filter approaches in other recommender systems.

\paragraph{Robustness}
Two types of users participate in our recommender system: the program managers and the prospective students. Changes in program and especially course description data can lead to corrupted results. We suggest the validation of the system to detect irrelevant keywords (see also Section \ref{subsec:programDataProcessing}). Such occurrences could also be detected on a continuous basis by including a corresponding question in a short user survey after having obtained a recommendation.

\paragraph{Bias and Privacy Protection}
We consider our system as practically unbiased since recommendations for distinct users are independent \citep{chen2021bias}. Popular programs - among students and in the provided ranking - are solely presented based on the user's interest topic selection. To ensure privacy protection \citep{Jeckmans2013}, neither the users' interests nor the recommended programs are shared with third parties.

\paragraph{Personalization}

To better understand the grade of personalization provided by our system, we quantify the dissimilarity of the 65 recommendations that our system generated for the participants using cosine similarity. Let $M^{R,P}\in \{0,1\}^{|R|\times |P|}$ be the binary experiment recommendation matrix with $M^{R,P}_{u,p}=1$ if program $p$ was included in the recommendation for user $u$, and $M^{R,P}_{u,p}=0$ otherwise. Then the cosine similarity between two recommendations (rows in $M^{R,P}$) be denoted by $s_{u,u'}$. It will always be in the interval $[0,1]$ since the underlying matrix is binary. The personalization score is calculated as $1-\overline{s}$ where $\overline{s}$ is the average similarity; i.e., $\overline{s}=1/\cdot\sum_{u<u'\in \{1,...,|R|\}}s_{u,u'}$. In our case study, we observe a high personalization score of 0.77 for program recommendation. The corresponding college personalization score is lower (0.48), which we attribute to the low number of available colleges (7). The complete experiment recommendation matrix is given in Appendix B.

Furthermore, the number of unique program recommendations was 59 (91.0\%) and the number of unique recommended colleges was 25 (38.5\%). More detailed, there were 9 matching program recommendation on the program level and 126 on the college level, respectively. At most 4 colleges were recommended to a single user. In 4 cases (6.2\%), the recommended programs were all located at a single college. 

\paragraph{Trustworthiness}
It is crucial for recommender systems to ensure that the user trusts the system~\citep{wang2023trustworthy}. We see two categories of \sys ~users: prospective students and educational program managers. For the former group, we use the coverage argument which is also used to support system fairness. To better understand the reception of our system by future students, we refer to survey results in the next section. In Question 6, over 94\% of the users stated that they would consider using \sys ~in the future. We claim that future students seeing their interests reflected (98.1\%; see Question 2) will create confidence.

\subsubsection{User Feedback}\label{subsubsec{UserFeedbackEval}}

\begin{figure}[htp]\centering
	\includegraphics[width=0.75\textwidth]{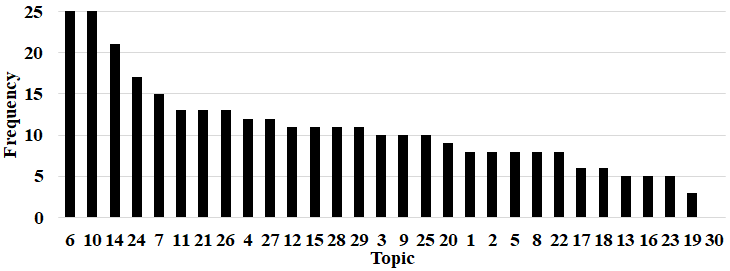}
    \hspace{6mm}
    \includegraphics[width=0.15\textwidth]{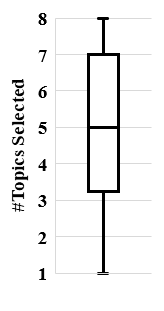}
	\caption{Popularity of topics among participants in our experiment (left), and the distribution of the number of interest topics selected by the participants (right). See Appendix for topics.}
	\label{fig:ClusterSelectionFreq}
\end{figure}

The most popular topics were picked by 25 participants (39\%). Only one topic (topic 30) was never selected in our experiment. On average, each topic was selected by 10.6 participants (16\%). The distribution of how many times a cluster was selected is presented in Figure~\ref{fig:ClusterSelectionFreq} (right). On average, students picked 4.9 topics.

For the current program metric, we need to keep in mind that not everyone chose their major based on their interests.
\begin{table}[htp]\caption{Results for the user survey binary questions addressing validity of the \sys ~decision support system (in percent).} \label{tab:surveyResultsBinary}
 \begin{tabular}{c lc cc c}
    \toprule
    \textbf{\#} & \textbf{Question} && \textbf{Yes} && \textbf{No}\\
    \midrule
    1 & ``My current program was recommended'' && 46.3 && 53.7\\
    \cmidrule{1-2}\cmidrule{4-4}\cmidrule{6-6}
    2 & ``One or more of the recommended programs matched my interests'' && 98.1 && 1.9\\
    \cmidrule{1-2}\cmidrule{4-4}\cmidrule{6-6}
    3 & ``There is an unexpected program that I did not consider yet'' && 74.5 && 25.5\\
    \cmidrule{1-2}\cmidrule{4-4}\cmidrule{6-6}
    4 & ``A program that I would consider (besides the one I'm enrolled in) && 57.4 && 42.6\\
     & is not recommended'' && && \\
    \bottomrule
 \end{tabular}
\end{table}
We show the user feedback regarding our binary questions in Table~\ref{tab:surveyResultsBinary}. 75\% of the students stated that there was an unexpected program suggested that they didn't consider yet. Moreover, more than 98\% indicated that one or more of the recommended programs matched their interests. It could be there are programs of interest among the unexpected programs. This would mean our method allows for serendipity in program recommendation.

Furthermore, we asked the experiment users questions concerning usefulness and usability of the system. Table \ref{tab:surveyResultsLikert} shoes these survey results that we obtained using a Likert scale. It can be seen that over 96\% of participant agreed that the tool can be useful. Similarly, more than 94\% would use consider using the system in the future. However, one user strongly disagreed to consider this support. The presented topic wordclouds were meaningful for about 93\% of the students. Even though 85\% did find it easy to select topics that match their interests, 13\% had a neutral feeling about this task. All participants stated that overall using this tool was easy. 

\begin{table}[htp]
\adjustbox{max width=\textwidth}{
 \begin{tabular}{c l cc cc cc cc cc c}
  \toprule
  \textbf{\#} & \textbf{Question} & &{\textbf{Strongly}} && {\textbf{Agree}} &&{\textbf{Neutral}} && {\textbf{Disagree}} &&{\textbf{Strongly}}\\
   & & &{\textbf{Agree}} && && && &&{\textbf{Disagree}}\\
  \midrule
  5 &``I think this tool can be useful to && 53.7 && 42.6 && 1.9 && 1.9 && 0.0\\
   & select a major.'' &&  &&  &&  &&  && \\
  \cmidrule{1-2}\cmidrule{4-4}\cmidrule{6-6}\cmidrule{8-8}\cmidrule{10-10}\cmidrule{12-12}
  6 & ``I would consider using this tool && 50.0 && 44.4 && 3.7 && 0.0 && 1.9\\
    & in addition to existing resources.'' &&  &&  &&  &&  && \\
  \cmidrule{1-2}\cmidrule{4-4}\cmidrule{6-6}\cmidrule{8-8}\cmidrule{10-10}\cmidrule{12-12}
  7 & ``I think the interest word clouds && 31.5 && 61.1 && 7.4 && 0.0 && 0.0\\
   & were mostly meaningful.'' &&  &&  &&  &&  && \\
  \cmidrule{1-2}\cmidrule{4-4}\cmidrule{6-6}\cmidrule{8-8}\cmidrule{10-10}\cmidrule{12-12}
  8 & ``It was easy for me to select && 50.0 && 35.2 && 13.0 && 1.9 && 0.0\\
   & interesting word clouds.'' &&  &&  &&  &&  && \\
  \cmidrule{1-2}\cmidrule{4-4}\cmidrule{6-6}\cmidrule{8-8}\cmidrule{10-10}\cmidrule{12-12}
  9 & ``Overall, it was easy to use this tool.'' && 77.8 && 22.2 && 0.0 && 0.0 && 0.0\\
  \bottomrule
  \end{tabular}
  }
\caption{Results for the user survey questions addressing usefulness and usability of the \sys ~decision support system (in percent).} \label{tab:surveyResultsLikert}
\end{table}

We asked for suggestions for improvement of this tool and all user comments are listed in Figure~\ref{fig:userComments}.
{\def\arraystretch{0.62}
\begin{table}[htp] \caption{All user comments after obtaining recommendations by \sys.} \label{fig:userComments}
 \begin{tabular}{c p{0.91\textwidth}}
  \toprule
  \textbf{\#} & \textbf{Comment}\\
  \midrule
  1 & ``Really awesome! Maybe do a blind study without people knowing this is a major test?''\\
  \midrule
  2 & ``This tool was a very interesting way to map interests to specific majors.''\\
  \midrule
  3 & ``Excluding the name of majors from clusters might decrease bias.''\\
  \midrule
  4 & ``Awesome!''\\
  \midrule
  5 & ``Maybe a step further and could recommend potential jobs/internships :) thought this was pretty snazzy.''\\
  \midrule
  6 & ``I think this tool is far more beneficial to students who have not declared a major yet, since it provides many different options (helpful for someone searching for a major idea). Overall a very cool idea and tool.''\\
  \midrule
  7 & ``As an art and design major I didn't see many clusters that described my particular program, maybe expanding those clusters to be sure they focus in on each major that is offered.''\\
  \midrule
  8 & ``Look into legibility research of different typefaces for fonts of word clusters.''\\
  \midrule
  9 & ``Maybe organize the word clusters so that they are spaced a little further apart.''\\
  \midrule
  10 & ``Consider using weighting for the different word clouds.''\\
  \midrule
  11 & ``Great work!''\\
  \midrule
  12 & ``I liked it.''\\
  \midrule
  13 & ``Include images in the clouds.''\\
  \midrule
  14 & ``None, this is a fantastic project!''\\
  \midrule
  15 & ``I think it was very easy to use and found it very interesting to see what I found interesting. I wish I could have seen this when I first applied to [the university].''\\
  \midrule
  16 & ``My of an explanation about how the clusters were derived and the relationship between them (but maybe that spoils too much of the point of this project).''\\
  \midrule
  17 & ``Seems like a great tool to get starting points for recommendations. If anything, having multiple rounds could be interesting, to further narrow down a selection.''\\
  \midrule
  18 & ``I think the words should be more theoretical it was very literal explanations and I could tell what the clusters would recommend to me.''\\
  \bottomrule
  \end{tabular}
\end{table}
}

Users provided a range of feedback on our tool, expressing both praise and suggestions for improvement, as shown in Table \ref{fig:userComments}. Many users found the tool's concept intriguing and innovative, particularly in its ability to map personal interests to specific majors. Positive sentiments were succinctly shared with comments like ``Awesome!'' and ``Great work!'', highlighting the overall appreciation for the tool's value and effectiveness. Users also praised the tool's potential to assist undecided students by presenting them with a diverse range of major options, making it especially beneficial for those who haven't declared a major yet. Some even expressed a sense of nostalgia, wishing they had access to such a tool when they first applied to the university.

However, among the areas for improvement identified by users, a notable suggestion was to conduct a blind study, which could potentially enhance the tool's credibility and effectiveness by removing any bias associated with participants knowing it’s a major test. Some users recommended refining the presentation of the word clusters for better legibility, including exploring different typefaces and adjusting the spacing between clusters to enhance readability and the user experience. The concept of incorporating images within the word clouds was also suggested, adding a visual element that could enrich the tool’s engagement.

Additionally, users proposed ways to improve the accuracy of the recommendations. One user recommended removing major names from clusters to reduce potential bias, while another suggested using weighted word clouds to better reflect the significance of certain interests within each major. Some users also expressed a desire for a more detailed explanation of how the clusters were derived and their relationships, though this request was balanced with the acknowledgment that too much detail might compromise the intended user experience of exploration and discovery.

Furthermore, users shared ideas for expanding the functionality beyond major recommendations. The idea of suggesting potential jobs and internships based on the identified interests was praised for its innovation and its potential to offer even more value to users. Some users also suggested introducing multiple rounds of recommendations to further narrow down options, which could help refine choices and provide a more tailored selection process.

In summary, the user feedback highlighted a balance of appreciation for the system’s unique concept and suggestions for improvement across various aspects. Users recognized its potential value for both undecided students and those seeking more specialized recommendations. The suggestions ranged from technical adjustments to more sophisticated features, all aimed at making the tool even more effective, engaging, and valuable to its users.

\section{Conclusion}\label{sec:conclusion}
In this paper, we presented a novel information system that supports prospective and current students in making informed decisions when choosing a study program. The system is applicable across a wide range of educational settings and can assist educational institutions in positively impacting student transfers, dropout rates, and overall satisfaction. Our tool leverages extensive course description data and applies advanced machine learning techniques to generate keyword clusters representing interest topics. Students match their personal interests with one or more of these suggested topics, which are then used to create a ranked list of the most relevant study programs, assisting students in their final decision-making process. \ale{While our approach is the first to incorporate program data at the course level, it relies on course descriptions being both well-written and accurate.}

We implemented and evaluated the system at a large polytechnic university. In a qualitative study involving more than 60 personalized rankings, we found that users highly valued the recommendations, considering them meaningful and helpful. Our quantitative analysis demonstrated how to configure the system to ensure fairness, achieving 98\% program coverage while maintaining a high personalization score of 0.77. Additionally, students reported experiencing both serendipity and trustworthiness in the recommendations. We are currently focused on integrating our system into existing student support services to enhance the overall program selection process.

\ale{For future research, we recommend extending our system to related areas, such as vocational and professional training. Additionally, integrating offerings from multiple universities, minors, certificates, and study abroad opportunities could further benefit students. Providing recommendations at the course level could also be an interesting feature. Another key consideration is exploring strategies to increase user control—such as allowing users to exclude certain topics or programs from recommendations based on factors like admission exam scores. Finally, we believe the accuracy of the system could be enhanced by incorporating more detailed course data, including syllabi, class activities, and course materials.}

\section*{Acknowledgements}
We would like to thank Daniel Waldorf for his valuable feedback and for naming the knowledge map. We are also grateful to all the users who participated in our experiments.

\section*{Declarations}
The authors have no competing interests to declare that are relevant to the content of this article.

\begin{appendices}

\subsection*{A. All 30 word clouds representing the interest topics for the example in Section~\ref{subsec:realRecommendation}.}

\begin{figure}[htp]\centering
  \includegraphics[width=0.69\textwidth]{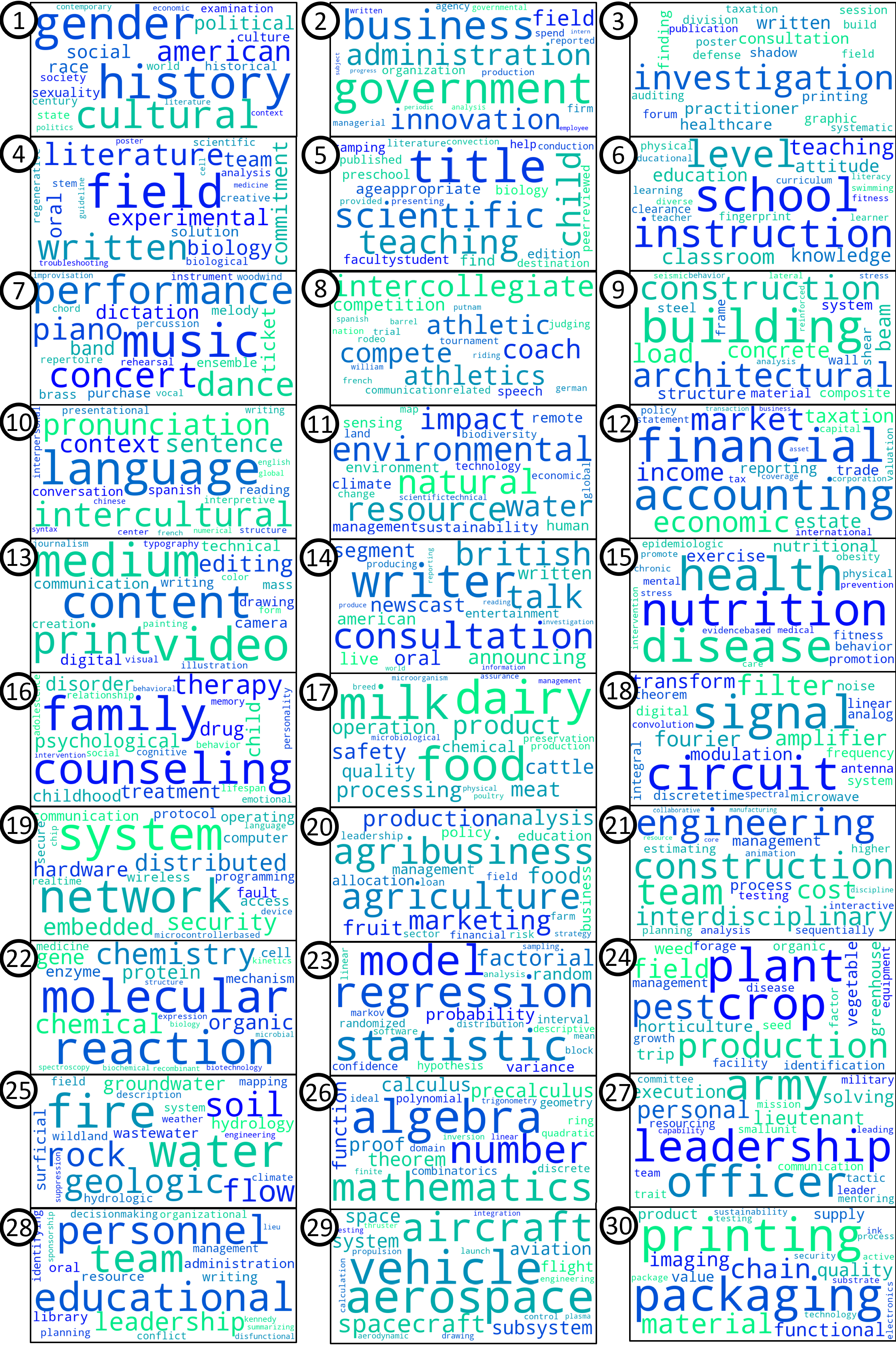}  
 \caption{All interest topics represented by word clouds. 25 (out of 30=$h$) that were not selected by the user in Section~\ref{subsec:realRecommendation}: \#1, \#3-18, \#20, \#22, \#24-29.}
 \label{fig:allWordclouds}
\end{figure}

\subsection*{B. The complete recommendation data for the case study in Section~\ref{sec:caseStudy}.}

\begin{figure}[htp]\centering
 \blind{
 \includegraphics[width=0.97\textwidth]{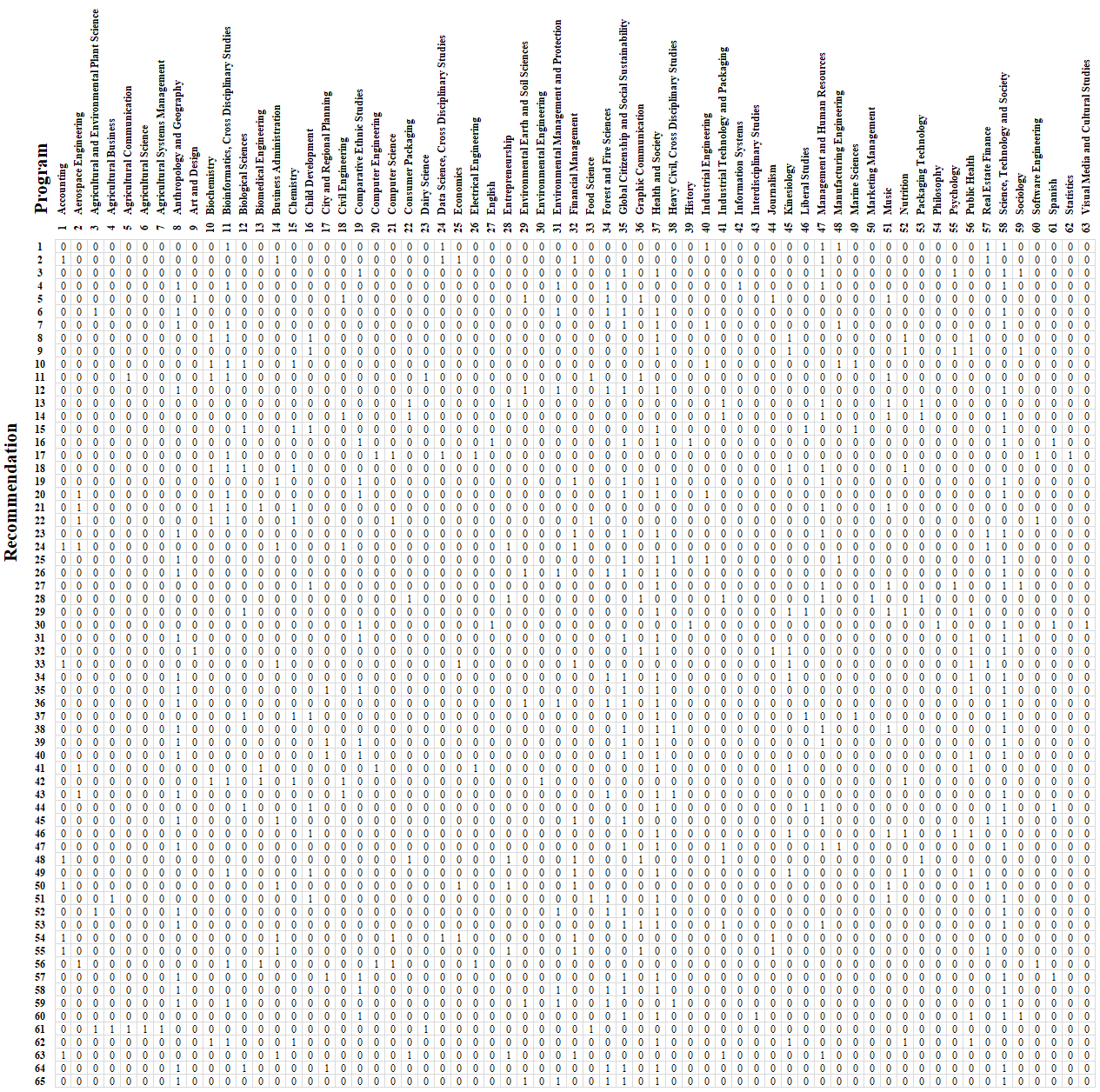}
 }
 \caption{The colleges recommended to the 65 experiment users, encoded in the binary program recommendation matrix $M^{R,P}$.}
 \label{fig:recommendationsProgramEncoded}
\end{figure}

\begin{figure}[htp]\centering
 \vspace{0mm}
 \blind{
 \includegraphics[width=0.25\textwidth]{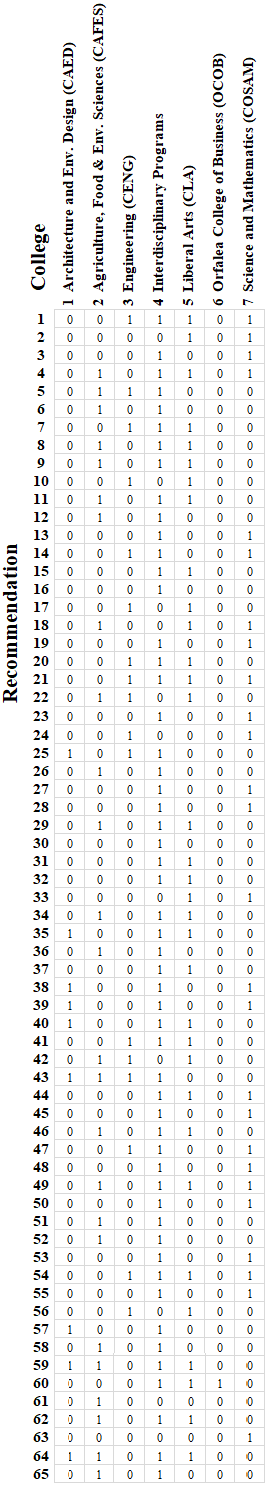}
 \hspace{12mm}\includegraphics[width=0.6\textwidth]{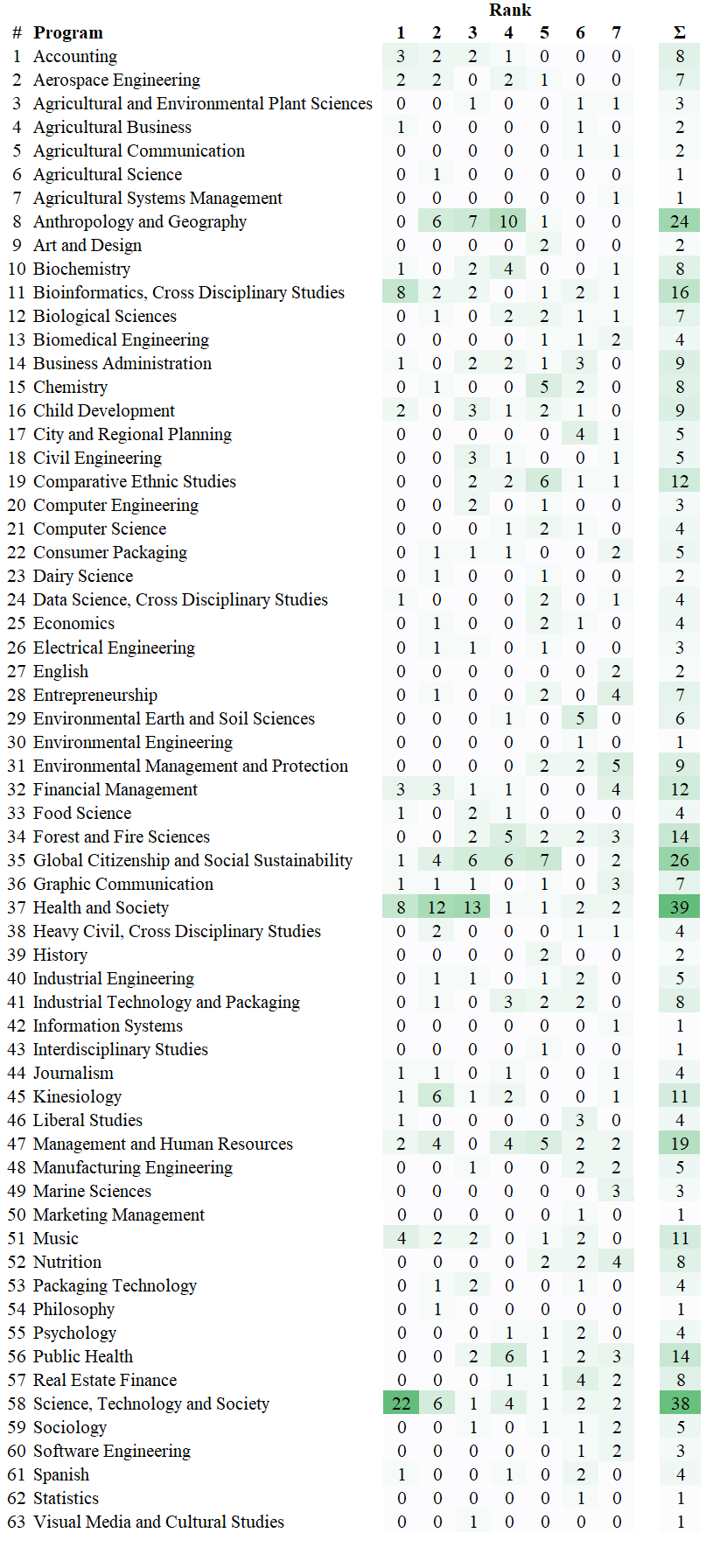}
 }
 \caption{The colleges recommended to the 65 experiment users, encoded in the binary college recommendation matrix $M^{R,C}$ (left), and program recommendation rank analysis (right).}
 \label{fig:recommendationsCollegeEncodedAndProgramRanks}
\end{figure}

\begin{table}[htp]\centering
 \vspace{0mm}
 \blind{
 \includegraphics[width=0.7\textwidth]{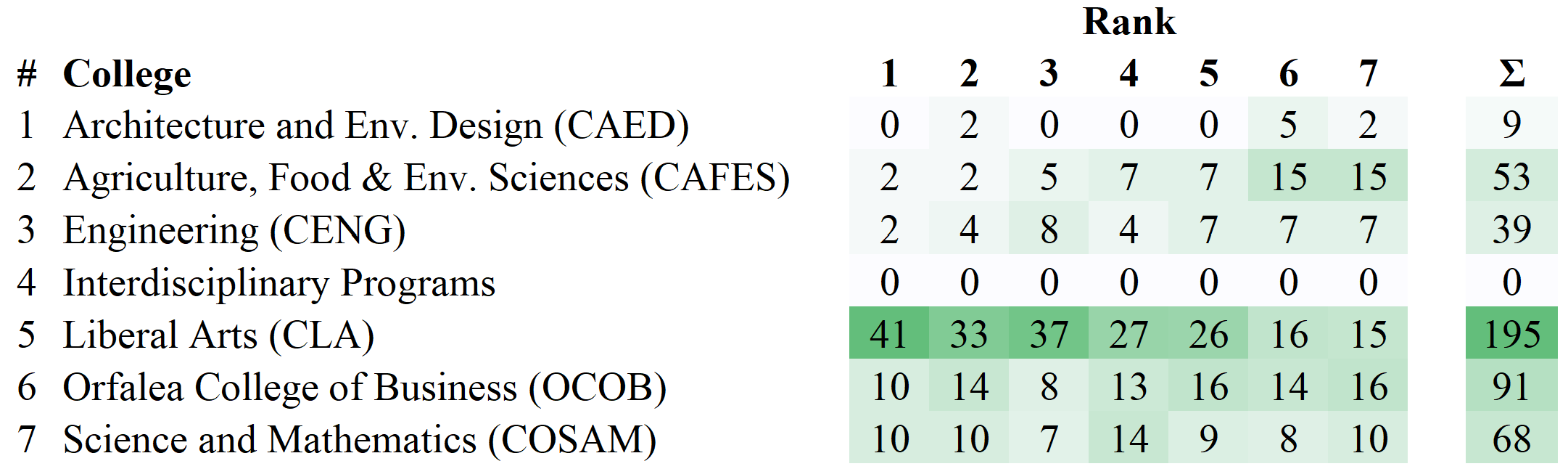}
 \caption{College recommendation rank analysis for all 65 experiment recommendations.}
 }
 \label{tab:recommendationAnalysisCollege}
\end{table}

\ale{
\subsection*{C. The complete PIS data for the case study in Section~\ref{sec:caseStudy}.}
}

\addtolength{\tabcolsep}{-0.5em}

\begin{table}[htp]
 \adjustbox{width=\textwidth}{
  \begin{tabular}{cc cccccccccc cccccccccc cccccccccc c}
   \toprule
\# &{\bf Program} & {\bf 1} & {\bf 2} & {\bf 3} & {\bf 4} & {\bf 5} & {\bf 6} & {\bf 7} & {\bf 8} & {\bf 9} & {\bf 10} & {\bf 11} & {\bf 12} & {\bf 13} & {\bf 14} & {\bf 15} & {\bf 16} & {\bf 17} & {\bf 18} & {\bf 19} & {\bf 20} & {\bf 21} & {\bf 22} & {\bf 23} & {\bf 24} & {\bf 25} & {\bf 26} & {\bf 27} & {\bf 28} & {\bf 29} & {\bf 30} & $|p|$ \\
\midrule
1 & Aerospace Engineering & 10 & 49 & 0 & 59 & 4 & 14 & 9 & 0 & 100 & 15 & 17 & 5 & 5 & 2 & 16 & 5 & 13 & 44 & 35 & 52 & 102 & 9 & 73 & 17 & 67 & 31 & 17 & 9 & 258 & 16 & 95 \\
2 & Agricultural and Environ. Plant Sci. & 26 & 41 & 7 & 28 & 9 & 6 & 1 & 3 & 28 & 35 & 82 & 10 & 7 & 14 & 20 & 1 & 102 & 6 & 10 & 109 & 61 & 42 & 19 & 217 & 31 & 7 & 2 & 45 & 18 & 18 & 85 \\
3 & Agricultural Business & 24 & 36 & 0 & 17 & 3 & 7 & 1 & 0 & 18 & 4 & 51 & 32 & 4 & 13 & 21 & 2 & 95 & 2 & 5 & 104 & 34 & 15 & 22 & 66 & 23 & 3 & 7 & 18 & 6 & 17 & 45 \\
4 & Agricultural Communication & 23 & 28 & 0 & 19 & 7 & 7 & 0 & 0 & 16 & 11 & 43 & 18 & 39 & 8 & 17 & 6 & 69 & 8 & 19 & 80 & 23 & 15 & 18 & 39 & 15 & 7 & 24 & 21 & 4 & 12 & 44 \\
5 & Agricultural Science & 50 & 60 & 1 & 23 & 4 & 12 & 3 & 2 & 30 & 7 & 137 & 41 & 17 & 18 & 36 & 12 & 197 & 8 & 24 & 177 & 83 & 18 & 24 & 177 & 63 & 5 & 18 & 64 & 18 & 29 & 107 \\
6 & Agricultural Systems Mngmt & 24 & 39 & 6 & 45 & 2 & 13 & 2 & 0 & 46 & 9 & 105 & 21 & 18 & 10 & 21 & 8 & 125 & 11 & 22 & 103 & 90 & 23 & 43 & 62 & 65 & 5 & 16 & 37 & 44 & 45 & 81 \\
7 & Animal Science & 36 & 108 & 5 & 84 & 19 & 34 & 6 & 9 & 86 & 27 & 180 & 52 & 5 & 19 & 111 & 49 & 300 & 20 & 23 & 238 & 140 & 159 & 56 & 174 & 43 & 19 & 10 & 96 & 43 & 48 & 202 \\
8 & Anthropology and Geography & 133 & 26 & 8 & 37 & 17 & 20 & 1 & 2 & 28 & 30 & 247 & 36 & 18 & 20 & 30 & 38 & 39 & 5 & 10 & 46 & 49 & 12 & 24 & 46 & 39 & 1 & 2 & 24 & 5 & 21 & 65 \\
9 & Architectural Engineering & 5 & 24 & 0 & 27 & 0 & 6 & 1 & 0 & 85 & 4 & 6 & 2 & 10 & 3 & 9 & 3 & 8 & 14 & 14 & 24 & 54 & 3 & 40 & 2 & 24 & 13 & 5 & 1 & 34 & 4 & 49 \\
10 & Architecture & 17 & 3 & 0 & 3 & 0 & 5 & 0 & 0 & 71 & 2 & 16 & 1 & 17 & 3 & 6 & 6 & 9 & 2 & 5 & 8 & 15 & 1 & 2 & 3 & 8 & 3 & 5 & 5 & 15 & 3 & 37 \\
11 & Art and Design & 134 & 28 & 10 & 23 & 1 & 9 & 2 & 4 & 14 & 14 & 14 & 8 & 136 & 15 & 4 & 18 & 16 & 14 & 10 & 24 & 49 & 7 & 8 & 14 & 1 & 0 & 18 & 11 & 14 & 16 & 88 \\
12 & Biochemistry & 8 & 30 & 4 & 69 & 30 & 37 & 0 & 0 & 43 & 15 & 15 & 3 & 8 & 1 & 23 & 8 & 37 & 7 & 5 & 31 & 36 & 179 & 31 & 22 & 16 & 13 & 1 & 0 & 23 & 14 & 76 \\
13 & Biological Sciences & 65 & 51 & 20 & 143 & 71 & 65 & 0 & 0 & 65 & 32 & 211 & 16 & 14 & 31 & 72 & 43 & 47 & 15 & 23 & 79 & 90 & 156 & 78 & 71 & 52 & 27 & 13 & 36 & 35 & 39 & 142 \\
14 & Biomedical Engineering & 13 & 57 & 1 & 77 & 7 & 13 & 0 & 0 & 92 & 20 & 19 & 6 & 16 & 3 & 21 & 11 & 27 & 73 & 33 & 55 & 113 & 73 & 87 & 13 & 53 & 58 & 10 & 10 & 78 & 53 & 98 \\
15 & BioResource and Agricultural Eng. & 19 & 22 & 3 & 38 & 5 & 7 & 0 & 0 & 37 & 10 & 59 & 9 & 11 & 4 & 13 & 4 & 38 & 18 & 11 & 32 & 64 & 25 & 35 & 23 & 62 & 17 & 7 & 8 & 56 & 17 & 63 \\
16 & Business Administration & 27 & 37 & 2 & 15 & 1 & 5 & 2 & 0 & 18 & 11 & 25 & 74 & 5 & 13 & 6 & 10 & 12 & 1 & 6 & 57 & 25 & 2 & 26 & 11 & 1 & 1 & 10 & 16 & 7 & 19 & 33 \\
17 & Chemistry & 8 & 28 & 4 & 62 & 32 & 37 & 0 & 0 & 33 & 16 & 13 & 3 & 8 & 1 & 17 & 7 & 35 & 11 & 16 & 29 & 36 & 138 & 45 & 16 & 17 & 22 & 2 & 0 & 26 & 12 & 69 \\
18 & Child Development & 27 & 5 & 1 & 18 & 24 & 22 & 0 & 0 & 15 & 5 & 13 & 3 & 6 & 2 & 21 & 88 & 3 & 3 & 1 & 5 & 3 & 13 & 11 & 0 & 0 & 2 & 2 & 5 & 0 & 1 & 39 \\
19 & City and Regional Planning & 192 & 64 & 14 & 48 & 8 & 15 & 1 & 0 & 56 & 24 & 295 & 58 & 38 & 27 & 21 & 67 & 32 & 10 & 13 & 80 & 101 & 1 & 63 & 33 & 46 & 10 & 20 & 63 & 15 & 25 & 124 \\
20 & Civil Engineering & 11 & 21 & 3 & 38 & 4 & 7 & 0 & 0 & 66 & 9 & 55 & 5 & 9 & 8 & 8 & 10 & 20 & 20 & 13 & 29 & 97 & 22 & 41 & 13 & 121 & 34 & 5 & 14 & 65 & 9 & 70 \\
21 & Communication Studies & 45 & 10 & 2 & 11 & 1 & 4 & 0 & 1 & 10 & 5 & 12 & 4 & 16 & 4 & 2 & 13 & 1 & 1 & 10 & 9 & 9 & 0 & 14 & 1 & 2 & 1 & 15 & 10 & 0 & 4 & 26 \\
22 & Comparative Ethnic Studies & 580 & 58 & 15 & 58 & 22 & 39 & 16 & 16 & 56 & 70 & 191 & 93 & 28 & 85 & 29 & 101 & 16 & 5 & 16 & 74 & 67 & 1 & 49 & 15 & 36 & 2 & 10 & 25 & 13 & 17 & 156 \\
23 & Computer Engineering & 14 & 45 & 4 & 51 & 4 & 13 & 5 & 0 & 59 & 19 & 33 & 4 & 14 & 5 & 15 & 10 & 35 & 93 & 88 & 56 & 88 & 19 & 70 & 23 & 30 & 52 & 11 & 16 & 49 & 63 & 83 \\
24 & Computer Science & 31 & 60 & 3 & 86 & 16 & 38 & 11 & 2 & 105 & 45 & 74 & 21 & 51 & 19 & 10 & 23 & 57 & 86 & 297 & 93 & 173 & 56 & 170 & 41 & 65 & 58 & 31 & 40 & 113 & 122 & 176 \\
25 & Construction Management & 11 & 25 & 0 & 18 & 4 & 9 & 2 & 0 & 88 & 4 & 37 & 20 & 7 & 4 & 8 & 6 & 18 & 4 & 4 & 37 & 95 & 6 & 25 & 12 & 19 & 5 & 7 & 15 & 17 & 15 & 54 \\
26 & Dairy Science & 12 & 35 & 3 & 25 & 6 & 14 & 3 & 0 & 29 & 12 & 50 & 28 & 14 & 10 & 41 & 11 & 175 & 7 & 11 & 121 & 48 & 59 & 27 & 88 & 12 & 8 & 8 & 39 & 15 & 18 & 89 \\
27 & Economics & 14 & 19 & 0 & 7 & 1 & 1 & 1 & 1 & 11 & 4 & 12 & 38 & 1 & 9 & 2 & 5 & 1 & 3 & 16 & 24 & 16 & 0 & 24 & 4 & 1 & 2 & 5 & 0 & 9 & 9 & 20 \\
28 & Electrical Engineering & 33 & 102 & 13 & 119 & 12 & 21 & 18 & 0 & 159 & 38 & 64 & 28 & 61 & 14 & 15 & 17 & 74 & 290 & 253 & 121 & 198 & 42 & 158 & 38 & 104 & 90 & 33 & 34 & 157 & 95 & 215 \\
29 & English & 86 & 12 & 5 & 26 & 18 & 19 & 0 & 6 & 7 & 94 & 6 & 4 & 26 & 31 & 0 & 6 & 2 & 1 & 18 & 11 & 10 & 0 & 10 & 4 & 0 & 0 & 1 & 16 & 1 & 1 & 60 \\
30 & Environmental Earth \& Soil Sci. & 136 & 82 & 14 & 69 & 21 & 34 & 3 & 2 & 85 & 44 & 473 & 44 & 41 & 28 & 54 & 47 & 161 & 17 & 31 & 196 & 177 & 58 & 66 & 226 & 300 & 8 & 22 & 111 & 52 & 51 & 221 \\
31 & Environmental Engineering & 13 & 18 & 1 & 33 & 2 & 11 & 1 & 0 & 33 & 10 & 61 & 7 & 9 & 9 & 22 & 10 & 30 & 13 & 6 & 31 & 52 & 49 & 44 & 21 & 65 & 17 & 6 & 11 & 45 & 13 & 60 \\
32 & Environ. Mngmt \& Protection & 154 & 93 & 14 & 76 & 26 & 35 & 3 & 2 & 79 & 37 & 503 & 55 & 39 & 25 & 52 & 51 & 171 & 16 & 21 & 220 & 185 & 67 & 73 & 245 & 266 & 6 & 17 & 126 & 41 & 54 & 224 \\
33 & Food Science & 21 & 27 & 1 & 28 & 10 & 10 & 3 & 0 & 20 & 12 & 42 & 8 & 5 & 12 & 28 & 12 & 160 & 4 & 1 & 87 & 33 & 46 & 20 & 39 & 14 & 6 & 1 & 10 & 15 & 43 & 66 \\
34 & Forest and Fire Sciences & 152 & 86 & 15 & 65 & 23 & 38 & 3 & 2 & 78 & 36 & 530 & 54 & 41 & 27 & 60 & 53 & 186 & 17 & 22 & 218 & 198 & 61 & 64 & 261 & 311 & 6 & 11 & 141 & 47 & 51 & 223 \\
35 & General Engineering & 6 & 24 & 0 & 40 & 9 & 5 & 2 & 0 & 44 & 21 & 8 & 2 & 4 & 0 & 6 & 5 & 11 & 13 & 12 & 29 & 61 & 47 & 44 & 12 & 23 & 19 & 10 & 4 & 35 & 14 & 52 \\
36 & Graphic Communication & 14 & 42 & 50 & 25 & 1 & 3 & 2 & 0 & 20 & 13 & 64 & 25 & 96 & 11 & 7 & 8 & 51 & 25 & 40 & 76 & 67 & 12 & 31 & 41 & 3 & 11 & 46 & 44 & 13 & 101 & 67 \\
37 & History & 134 & 7 & 0 & 8 & 0 & 7 & 0 & 5 & 9 & 49 & 19 & 19 & 3 & 12 & 0 & 13 & 0 & 2 & 8 & 8 & 7 & 0 & 10 & 1 & 2 & 0 & 0 & 4 & 2 & 1 & 45 \\
38 & Industrial Engineering & 48 & 84 & 16 & 100 & 13 & 10 & 4 & 0 & 87 & 22 & 123 & 38 & 14 & 26 & 11 & 17 & 48 & 35 & 57 & 114 & 197 & 17 & 124 & 41 & 47 & 31 & 24 & 52 & 74 & 84 & 102 \\
39 & Industrial Tech. and Packaging & 4 & 9 & 1 & 5 & 0 & 5 & 1 & 0 & 4 & 5 & 20 & 8 & 3 & 8 & 3 & 1 & 23 & 2 & 5 & 17 & 14 & 5 & 14 & 11 & 1 & 2 & 5 & 13 & 6 & 37 & 24 \\
40 & Interdisciplinary Studies & 54 & 4 & 0 & 7 & 0 & 4 & 0 & 0 & 4 & 3 & 7 & 9 & 7 & 4 & 0 & 11 & 0 & 3 & 4 & 6 & 13 & 0 & 26 & 1 & 1 & 3 & 4 & 3 & 4 & 3 & 18 \\
41 & Journalism & 34 & 18 & 6 & 7 & 1 & 4 & 1 & 2 & 7 & 15 & 12 & 16 & 88 & 44 & 1 & 9 & 11 & 14 & 11 & 28 & 13 & 1 & 32 & 14 & 0 & 6 & 15 & 23 & 3 & 2 & 46 \\
42 & Kinesiology & 54 & 29 & 13 & 49 & 14 & 40 & 4 & 4 & 42 & 13 & 49 & 9 & 15 & 21 & 183 & 98 & 24 & 9 & 18 & 35 & 28 & 34 & 22 & 15 & 8 & 5 & 29 & 34 & 14 & 12 & 89 \\
43 & Landscape Architecture & 24 & 6 & 4 & 11 & 3 & 18 & 0 & 0 & 16 & 1 & 34 & 2 & 18 & 5 & 5 & 5 & 10 & 10 & 7 & 12 & 27 & 3 & 13 & 16 & 10 & 6 & 12 & 8 & 20 & 9 & 45 \\
44 & Liberal Arts and Engineering Studies & 35 & 43 & 17 & 38 & 2 & 11 & 9 & 3 & 50 & 17 & 41 & 8 & 35 & 11 & 6 & 8 & 33 & 33 & 64 & 52 & 102 & 3 & 66 & 23 & 31 & 22 & 16 & 22 & 50 & 63 & 82 \\
45 & Liberal Studies & 293 & 35 & 7 & 94 & 72 & 153 & 5 & 7 & 58 & 58 & 88 & 28 & 17 & 61 & 66 & 120 & 41 & 11 & 28 & 60 & 50 & 66 & 27 & 34 & 33 & 39 & 15 & 29 & 31 & 21 & 160 \\
46 & Manufacturing Engineering & 48 & 87 & 16 & 100 & 16 & 12 & 4 & 0 & 93 & 26 & 132 & 39 & 20 & 26 & 18 & 13 & 68 & 31 & 60 & 115 & 214 & 30 & 106 & 47 & 52 & 28 & 23 & 51 & 78 & 118 & 115 \\
47 & Marine Sciences & 49 & 40 & 22 & 102 & 53 & 48 & 2 & 0 & 44 & 34 & 171 & 14 & 15 & 35 & 17 & 22 & 33 & 10 & 20 & 57 & 86 & 90 & 72 & 40 & 48 & 23 & 8 & 28 & 34 & 24 & 118 \\
48 & Materials Engineering & 18 & 46 & 7 & 63 & 18 & 37 & 5 & 0 & 78 & 19 & 43 & 19 & 12 & 9 & 23 & 7 & 44 & 16 & 24 & 57 & 108 & 47 & 58 & 21 & 41 & 24 & 11 & 11 & 58 & 73 & 100 \\
49 & Mathematics & 7 & 40 & 3 & 42 & 2 & 18 & 4 & 1 & 51 & 22 & 12 & 11 & 3 & 7 & 8 & 3 & 17 & 59 & 40 & 49 & 60 & 8 & 119 & 8 & 19 & 106 & 10 & 9 & 35 & 22 & 108 \\
50 & Mechanical Engineering & 16 & 62 & 13 & 83 & 9 & 22 & 7 & 0 & 99 & 17 & 34 & 3 & 26 & 21 & 30 & 5 & 56 & 41 & 66 & 57 & 149 & 26 & 84 & 33 & 65 & 33 & 17 & 13 & 93 & 79 & 126 \\
51 & Microbiology & 16 & 40 & 7 & 74 & 20 & 8 & 1 & 0 & 44 & 15 & 43 & 5 & 4 & 8 & 40 & 7 & 90 & 12 & 15 & 68 & 41 & 146 & 51 & 58 & 19 & 13 & 6 & 7 & 28 & 23 & 94 \\
52 & Music & 22 & 14 & 9 & 27 & 7 & 10 & 170 & 0 & 8 & 8 & 1 & 0 & 8 & 15 & 0 & 2 & 2 & 5 & 1 & 7 & 5 & 3 & 5 & 7 & 0 & 1 & 0 & 3 & 0 & 2 & 60 \\
53 & Nutrition & 39 & 35 & 5 & 65 & 25 & 15 & 6 & 0 & 49 & 18 & 37 & 16 & 12 & 15 & 119 & 75 & 47 & 12 & 16 & 55 & 34 & 121 & 28 & 31 & 13 & 17 & 8 & 9 & 22 & 16 & 102 \\
54 & Philosophy & 43 & 9 & 2 & 13 & 3 & 8 & 0 & 8 & 10 & 2 & 22 & 4 & 1 & 4 & 6 & 15 & 2 & 0 & 1 & 12 & 10 & 1 & 8 & 3 & 1 & 2 & 1 & 2 & 4 & 16 & 38 \\
55 & Physics & 13 & 14 & 4 & 28 & 6 & 4 & 0 & 0 & 17 & 5 & 8 & 0 & 3 & 4 & 4 & 8 & 6 & 18 & 3 & 13 & 16 & 8 & 24 & 0 & 4 & 20 & 3 & 0 & 10 & 14 & 39 \\
56 & Political Science & 188 & 48 & 3 & 14 & 2 & 3 & 0 & 5 & 25 & 17 & 49 & 39 & 11 & 18 & 9 & 43 & 8 & 4 & 11 & 39 & 23 & 3 & 26 & 9 & 9 & 4 & 10 & 19 & 11 & 9 & 80 \\
57 & Public Health & 128 & 33 & 5 & 41 & 14 & 33 & 2 & 4 & 58 & 75 & 106 & 19 & 27 & 18 & 207 & 119 & 32 & 8 & 30 & 56 & 39 & 55 & 48 & 30 & 8 & 9 & 21 & 34 & 12 & 19 & 116 \\
58 & Psychology & 107 & 14 & 4 & 31 & 14 & 24 & 0 & 1 & 26 & 12 & 22 & 8 & 7 & 12 & 33 & 107 & 4 & 2 & 6 & 20 & 13 & 13 & 16 & 2 & 1 & 3 & 10 & 23 & 6 & 5 & 57 \\
59 & Recreation, Parks \& Tourism Admin. & 137 & 73 & 12 & 50 & 18 & 26 & 3 & 5 & 46 & 22 & 198 & 57 & 37 & 42 & 24 & 65 & 77 & 7 & 12 & 160 & 111 & 4 & 46 & 69 & 15 & 9 & 47 & 101 & 10 & 37 & 115 \\
60 & Sociology & 168 & 45 & 1 & 23 & 7 & 17 & 1 & 4 & 39 & 17 & 51 & 23 & 11 & 13 & 15 & 95 & 15 & 8 & 18 & 48 & 40 & 5 & 32 & 14 & 14 & 2 & 16 & 25 & 13 & 16 & 66 \\
61 & Software Engineering & 24 & 51 & 3 & 76 & 14 & 34 & 7 & 2 & 95 & 44 & 70 & 15 & 29 & 18 & 10 & 28 & 51 & 72 & 242 & 84 & 145 & 57 & 162 & 36 & 53 & 57 & 26 & 37 & 96 & 99 & 154 \\
62 & Spanish & 322 & 44 & 19 & 78 & 42 & 42 & 12 & 70 & 46 & 155 & 110 & 49 & 32 & 59 & 19 & 62 & 5 & 3 & 29 & 39 & 49 & 4 & 28 & 4 & 9 & 0 & 7 & 27 & 4 & 9 & 132 \\
63 & Statistics & 2 & 25 & 3 & 30 & 2 & 1 & 2 & 0 & 28 & 9 & 4 & 0 & 6 & 5 & 0 & 1 & 6 & 24 & 26 & 26 & 48 & 0 & 110 & 4 & 4 & 41 & 9 & 6 & 26 & 22 & 50 \\
64 & Theatre Arts & 22 & 18 & 9 & 18 & 3 & 5 & 16 & 1 & 26 & 11 & 4 & 2 & 17 & 24 & 4 & 5 & 11 & 2 & 5 & 15 & 21 & 5 & 9 & 9 & 2 & 0 & 3 & 8 & 2 & 11 & 43 \\
65 & Wine and Viticulture & 40 & 66 & 2 & 49 & 15 & 6 & 1 & 3 & 36 & 52 & 72 & 66 & 16 & 16 & 14 & 11 & 84 & 5 & 11 & 146 & 55 & 61 & 31 & 93 & 19 & 5 & 8 & 46 & 10 & 35 & 101 \\
66 & Accounting & 2 & 2 & 8 & 0 & 0 & 0 & 0 & 0 & 1 & 1 & 2 & 37 & 2 & 8 & 0 & 0 & 2 & 0 & 0 & 8 & 6 & 1 & 0 & 1 & 1 & 0 & 0 & 1 & 4 & 3 & 11 \\
67 & Consumer Packaging & 1 & 6 & 3 & 3 & 0 & 2 & 0 & 0 & 3 & 2 & 10 & 4 & 5 & 2 & 3 & 1 & 11 & 1 & 0 & 9 & 11 & 3 & 7 & 3 & 0 & 1 & 1 & 2 & 3 & 40 & 15 \\
68 & Entrepreneurship & 10 & 20 & 2 & 12 & 0 & 1 & 0 & 0 & 7 & 1 & 12 & 23 & 6 & 4 & 0 & 7 & 6 & 0 & 6 & 18 & 16 & 2 & 3 & 6 & 1 & 0 & 14 & 11 & 5 & 25 & 16 \\
69 & Ethics, Law and Social Justice & 85 & 8 & 1 & 8 & 0 & 1 & 0 & 3 & 11 & 2 & 11 & 12 & 5 & 5 & 4 & 20 & 0 & 2 & 0 & 10 & 15 & 4 & 9 & 1 & 1 & 1 & 0 & 1 & 5 & 6 & 34 \\
70 & Financial Management & 12 & 16 & 6 & 5 & 0 & 1 & 0 & 1 & 14 & 7 & 20 & 102 & 1 & 7 & 1 & 3 & 11 & 3 & 13 & 52 & 22 & 6 & 14 & 10 & 2 & 6 & 1 & 7 & 5 & 16 & 32 \\
71 & Global Citizenship \& Social Sustain. & 205 & 14 & 12 & 20 & 12 & 9 & 0 & 5 & 20 & 24 & 168 & 40 & 17 & 30 & 4 & 47 & 12 & 1 & 8 & 27 & 45 & 1 & 10 & 9 & 17 & 0 & 6 & 28 & 5 & 16 & 50 \\
72 & Health and Society & 156 & 8 & 14 & 37 & 23 & 20 & 0 & 2 & 21 & 18 & 154 & 13 & 16 & 17 & 55 & 89 & 17 & 2 & 7 & 29 & 41 & 8 & 7 & 11 & 10 & 3 & 7 & 20 & 9 & 18 & 43 \\
73 & Industrial Technology & 4 & 26 & 6 & 8 & 0 & 4 & 2 & 0 & 8 & 2 & 21 & 21 & 7 & 6 & 3 & 4 & 22 & 0 & 2 & 30 & 38 & 2 & 4 & 19 & 0 & 0 & 7 & 16 & 9 & 65 & 27 \\
74 & Information Systems & 2 & 12 & 0 & 3 & 0 & 3 & 1 & 0 & 5 & 2 & 11 & 8 & 3 & 1 & 0 & 1 & 7 & 1 & 10 & 20 & 20 & 0 & 6 & 7 & 2 & 0 & 4 & 11 & 6 & 8 & 12 \\
75 & Management and Human Resources & 5 & 22 & 0 & 8 & 0 & 6 & 7 & 0 & 7 & 2 & 28 & 12 & 0 & 1 & 0 & 5 & 7 & 0 & 0 & 23 & 23 & 1 & 3 & 11 & 0 & 1 & 11 & 21 & 2 & 25 & 14 \\
76 & Marketing Management & 3 & 14 & 2 & 6 & 0 & 2 & 0 & 0 & 3 & 0 & 3 & 2 & 5 & 5 & 0 & 3 & 6 & 3 & 3 & 26 & 8 & 0 & 6 & 3 & 0 & 0 & 3 & 2 & 5 & 9 & 12 \\
77 & Packaging Technology & 0 & 10 & 4 & 2 & 0 & 2 & 0 & 0 & 2 & 0 & 3 & 4 & 4 & 5 & 2 & 0 & 20 & 0 & 0 & 13 & 12 & 3 & 3 & 3 & 0 & 0 & 1 & 0 & 4 & 39 & 15 \\
78 & Quantitative Analysis & 4 & 23 & 0 & 20 & 0 & 1 & 1 & 0 & 21 & 4 & 5 & 12 & 1 & 3 & 1 & 1 & 4 & 35 & 18 & 31 & 26 & 0 & 73 & 2 & 3 & 46 & 3 & 2 & 13 & 10 & 41 \\
79 & Real Estate Finance & 8 & 9 & 2 & 4 & 0 & 0 & 0 & 1 & 10 & 2 & 27 & 28 & 1 & 2 & 1 & 1 & 8 & 2 & 3 & 17 & 16 & 2 & 9 & 11 & 5 & 1 & 2 & 8 & 2 & 6 & 14 \\
80 & Science, Technology and Society & 107 & 5 & 15 & 27 & 22 & 11 & 0 & 6 & 10 & 20 & 188 & 17 & 25 & 20 & 12 & 30 & 10 & 2 & 16 & 20 & 41 & 0 & 4 & 8 & 24 & 1 & 15 & 22 & 9 & 43 & 36 \\
81 & Visual Media and Cultural Studies & 141 & 25 & 7 & 29 & 10 & 5 & 16 & 10 & 19 & 22 & 11 & 3 & 24 & 25 & 0 & 12 & 8 & 4 & 9 & 24 & 28 & 0 & 18 & 9 & 2 & 0 & 3 & 2 & 3 & 8 & 53 \\
82 & Bioinformatics & 0 & 12 & 1 & 23 & 5 & 2 & 2 & 0 & 16 & 8 & 4 & 0 & 1 & 2 & 2 & 1 & 6 & 1 & 11 & 12 & 27 & 41 & 22 & 5 & 1 & 2 & 4 & 5 & 10 & 14 & 16 \\
83 & Data Science & 0 & 11 & 0 & 12 & 0 & 4 & 2 & 0 & 13 & 7 & 5 & 0 & 4 & 1 & 0 & 1 & 5 & 12 & 19 & 14 & 30 & 0 & 42 & 5 & 2 & 18 & 5 & 7 & 14 & 11 & 21 \\
84 & Heavy Civil & 2 & 5 & 0 & 6 & 0 & 0 & 1 & 0 & 40 & 0 & 14 & 3 & 3 & 5 & 0 & 0 & 8 & 0 & 2 & 12 & 55 & 0 & 7 & 16 & 10 & 0 & 0 & 9 & 12 & 7 & 19 \\
 \bottomrule
  \end{tabular}
  }
  \caption{The complete program interest scores and the numbers of program courses $|p|$ for the 84 programs and 30 clusters in the case study.}\label{tab:dataPIS}
\end{table}

\end{appendices}

\bibliography{topic}

\end{document}